\documentclass[11pt]{article}

\usepackage[final]{acl}

\usepackage{times}
\usepackage{latexsym}   
\usepackage{graphicx}
\usepackage{amsmath}
\usepackage{multirow} 
\usepackage{booktabs}
\usepackage{amssymb}
\usepackage{cleveref}
\usepackage{xcolor}
\usepackage[T1]{fontenc}

\usepackage[utf8]{inputenc}

\usepackage{microtype}

\usepackage{inconsolata}

\usepackage{graphicx}

\title{Knowledge Injection Exists in MoE? Exploring Expert-Aware Contrast Decoding in MoE for Mitigating LLMs' Hallucinations}

\author{
 \textbf{Xinyue Fang\textsuperscript{1}},
 \textbf{Zhiliang Tian\textsuperscript{1}}\thanks{Corresponding Author},
 \textbf{Zhen Huang\textsuperscript{1}}\footnotemark[1],
 \textbf{Ziyi Pan\textsuperscript{1}},
 \textbf{Zhihua Wen\textsuperscript{1}},
 \\
 \textbf{Xi Wang\textsuperscript{1},}
 \textbf{Quntian Fang\textsuperscript{1},}
 \textbf{Dongsheng Li \textsuperscript{1}}
\\
 \textsuperscript{1}College of Computer Science and Technology, National University of Defense Technology
\\
 \small{
\texttt{\{}
\href{mailto:fangxinyue@nudt.edu.cn}{fangxinyue},
\href{mailto:tianzhiliang@nudt.edu.cn}{tianzhiliang},
\href{mailto:huangzhen@nudt.edu.cn}{huangzhen},
\href{mailto:panziyi@nudt.edu.cn}{panziyi},
\href{mailto:zhwen@nudt.edu.cn}{zhwen},
\href{mailto:wangxi25@nudt.edu.cn}{wangxi25},
\href{mailto:fangquntian@nudt.edu.cn}{fangquntian},
\href{mailto:dsli@nudt.edu.cn}{dsli}
\texttt{\}@nudt.edu.cn}
 }
}

\begin{document}
\maketitle
\begin{abstract}
Existing LLM hallucination mitigation methods, including prompt engineering and model optimization, either hardly alter models' internal knowledge or have poor cross-domain generalization. Contrastive decoding mitigates hallucinations by using layer-wise differences in LLMs. However, prior studies only explore transformer-based models (e.g., GPT), ignoring other effective frameworks like mixture-of-experts (MoE) models. Since MoE alters the traditional transformer architecture, we conduct empirical studies to investigate whether similar layer-wise differences exist in MoEs. Our results show that they do not exist in MoE with shared experts; nevertheless, across different MoEs, higher layers exhibit distinct expert activation patterns between factual and non-factual outputs. Building on these, we propose \textbf{EAACD}, an expert-aware adaptive contrast decoding that uses expert differences in MoE's higher layers to mitigate hallucinations on QA tasks. EAACD splits high‑layer experts into a higher-reliability group and several lower-reliability groups based on their confidence and consistency. It contrasts the higher-reliability group's prediction with each lower-reliability group's prediction to calibrate the model's original predictions. To strengthen this contrast, EAACD amplifies hallucinations from lower-reliability experts via attention and masking to provide stronger negative references. EAACD outperforms all baselines on four datasets\footnote{Code is: \url{anonymous.4open.science/r/EAACD-D388/}}.

\end{abstract}

\section{Introduction}

Large language models (LLMs) show strong performance but suffer from hallucination, limiting their application \cite{fang2025zero,liu2025mitigating}. Existing mitigation methods mainly include: (1) \textit{Prompt engineering} uses task instructions to guide models to generate factual outputs \cite{mondal2024kam}. It is easy to apply, but cannot fundamentally alter the model’s internal knowledge \cite{cheng2025think}. (2) \textit{Model parameter optimization} fine-tunes LLMs with supplementary knowledge \cite{wang2022self}. These methods mitigate hallucinations by calibrating the model’s internal knowledge. But they lack domain generalization, and potential errors in fine-tuning data may exacerbate the hallucination \cite{iyer2022opt}.
\begin{figure*}[htp]
    \includegraphics[width=\textwidth]{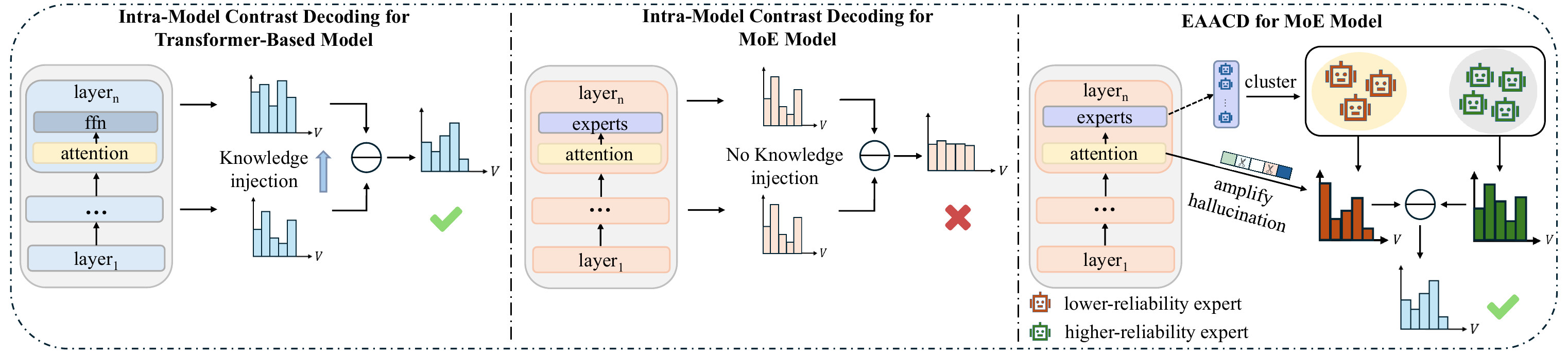}
    \caption{Overview of prior intra-model contrast decoding and EAACD (ours). Prior intra-model contrast decoding relies on ``knowledge injection'' in transformer-based models, limiting its applicability to MoE. Our EAACD leverages expert diversity within MoE layers to mitigate hallucinations effectively.}
    \label{fig:intro}
\end{figure*}

To improve domain generalization and reduce reliance on fine-tuning data, researchers propose contrastive decoding, which improves the output's accuracy by using less reliable outputs as negative references \cite{li2023contrastive}. This approach includes: (1) \textit{Inter‑model contrastive decoding} compares the original model's outputs with an unreliable model's outputs \cite{yang2024frustratingly}. These methods use the unreliable model’s outputs as negative references, removing such negative information from the original model’s outputs. However, if the unreliable model produces factual outputs, this may adversely affect the original model's predictions \cite{yin2025mirage}. (2) \textit{Intra‑model contrastive decoding}  leverages internal differences within a single model, contrasting logits between lower and higher layers to obtain next‑token probabilities \cite{yang2025improving}. Researchers \cite{chuang2023dola} find that transformer-based models (e.g., GPT) have a ``knowledge injection'' phenomenon where higher layers incorporate more factual knowledge during generation, while the lower layers store less factual knowledge. Following this, researchers develop several variants \cite{gera2023benefits,das2025entropy}, which use layer-wise differences to mitigate hallucinations without external resources.

Recently, the Mixture‑of‑Experts (MoE) architecture has become a popular approach for building large LLMs (e.g., DeepSeek-R1 \cite{guo2025deepseek}, GPT-4 \cite{achiam2023gpt}) due to its excellent performance. However, even advanced MoE models still suffer from hallucination \cite{su2025cartesianmoe}. MoE splits the model into multiple experts and dynamically routes each token to a small subset of them \cite{zhao2024hypermoe}. This design alters the model’s structure, making the existing intra‑model contrastive decoding for classical transformer-based models face challenges when mitigating hallucinations in MoE models. Because MoE models may not exhibit the same layer-wise differences as classical transformer-based models. 

To address these problems, we explore whether MoE models exhibit the ``knowledge injection'' found in classical transformer-based models. We find that this phenomenon depends on the MoE architecture: (1) \textbf{the ``knowledge injection'' only occurs in MoE without shared experts, but not in MoE with shared experts.} This suggests that the intra‑model contrastive decoding may only be effective in the MoE model without shared experts. Given this limitation, we explore whether we could use differences among experts for contrastive decoding. By examining expert activation patterns during factual versus non-factual outputs, we find: (2) \textbf{higher layers consistently show distinct expert activation patterns between factual and non-factual outputs across all MoE architectures.} This insight enables us to use higher-layer expert differences for intra‑model contrastive decoding to mitigate hallucinations in all MoEs.

In this paper, we propose an expert‑aware adaptive contrastive decoding strategy (\textbf{EAACD}) that separates higher-reliability experts from lower-reliability experts; and higher-reliability experts treat amplified hallucinations in lower-reliability experts as a negative reference to enhance factuality via contrastive decoding (Fig.~\ref{fig:intro}). Our empirical findings show that across MoE architectures, higher layers exhibit divergent expert activation when generating factual versus non-factual outputs. Since the router aims to select experts most suitable for specific scenarios, its differing preferences in factual and non-factual situations indicate varying expert capabilities in factual output generation. To use the inherent difference between experts for contrastive decoding, we split experts into one high-reliability and multiple low-reliability groups based on confidence and consistency. To ensure lower-reliability experts provide a valid negative reference during contrast, we amplify hallucinations in low-reliability experts via attention and masking mechanisms. Finally, our contrast decoding module adaptively penalizes lower-reliability predictions based on their differences from the higher-reliability prediction during contrast, and calibrates original predictions using contrast results to mitigate hallucinations on QA tasks. 

Our main contributions are: (1) To our knowledge, we are the first to explore the ``knowledge injection'' in MoE models and expert activation differences between factual and non-factual outputs. (2) We reveal how ``knowledge injection'' and expert activation differences during factual versus non-factual outputs relate to the MoE's architecture. (3) We propose an expert-aware adaptive contrastive decoding method to mitigate MoE models' hallucinations without external resources. (4) Our method is SOTA across four datasets.

\section{Related Work}
\label{relat}
\textbf{Hallucination Mitigation in LLMs} aims to reduce factual inconsistencies in LLMs outputs.\cite{si2022prompting}. Existing mitigation methods: \textbf{(1) Prompt Engineering} improves the model’s factuality by including a few examples in the prompt \cite{brown2020language}. Chain‑of‑Thought \cite{mondal2024kam} prompting guided models through intermediate steps to boost factuality. \citet{ji2023towards} proposed an iterative self-reflection incorporating prior outputs for correction. \textbf{(2) Optimizing Model Parameters} aligns the model’s internal knowledge with facts \cite{wang2022self}. \textit{Supervised Fine-Tuning} (SFT) helps LLMs learn task-specific features from labeled data and improve accuracy \cite{iyer2022opt,shi2023hallucination}. \textit{Model Editing} removes hallucinations via localization and intervention \cite{zheng2023can,zhang2024adversarial}. 

\textbf{Contrastive decoding mitigates hallucinations} by comparing outputs between models or different parts of the same model \cite{li2023contrastive,chuang2023dola}, mainly divided into two types: \textbf{(1) Inter-model contrastive decoding} compares outputs from a large model with a less accurate model \cite{chen2024freeze}.  \citet{li2023contrastive} used a smaller model to calibrate the larger model's output. \citet{zhang2025alleviating} fine-tuned the original model on hallucinated data to construct a less factual model as a negative comparator for the original model. 
\textbf{(2) Intra-model contrastive decoding} mitigates hallucinations by leveraging internal discrepancies in the model \cite{wu2025improve,shi2024unchosen}. DoLa \cite{chuang2023dola} contrasted the model’s final-layer prediction with earlier layers to highlight reliable outputs. \citet{das2025entropy} selected lower layers based on layer-wise entropy, improving DoLa’s performance. Context-Aware Decoding \cite{shi2024trusting} compared context-aware with context-agnostic outputs to mitigate hallucinations. 

\textbf{Mixture-of-Experts (MoE)} \cite{li2025uni} enables LLMs to scale model capacity while maintaining efficiency. MoE architectures are categorized as: \textbf{(1) MoE without shared experts} fully decouples experts \cite{jiang2024mixtral}. Mixtral 8x7B \cite{jiang2024mixtral} used equally sized, independent experts and activated top-2 during inference. LLaMA‑MoE \cite{zhu2024llama} employed non-overlapping partitions for better generalization. \textbf{(2) MoE with shared experts} has shared experts always activated, while others are routed \cite{zhao2024hypermoe}. DeepSeekMoE \cite{dai2024deepseekmoe} and XMoE \cite{yang2024xmoe} split experts into sub-experts and activated top-K at inference. Cartesian‑MoE \cite{su2025cartesianmoe} merged shared and routed experts to enlarge compositional capacity. 

\section{Empirical Study}
We conduct an empirical study to explore general differences in  MoE models that enable intra‑model contrast decoding for hallucination mitigation. It has two key experiments. First, we examine whether MoE models with different architectures exhibit the ``knowledge injection'' phenomenon observed in classical transformer-based models. Results show ``knowledge injection'' only occurs in MoE without shared experts, but not in MoE with shared experts. Therefore, we turn to explore more general differences between factual and non-factual outputs across all MoE models.
\begin{figure}[htp]
    \centering
    \includegraphics[width=\columnwidth,height=0.26\textheight]{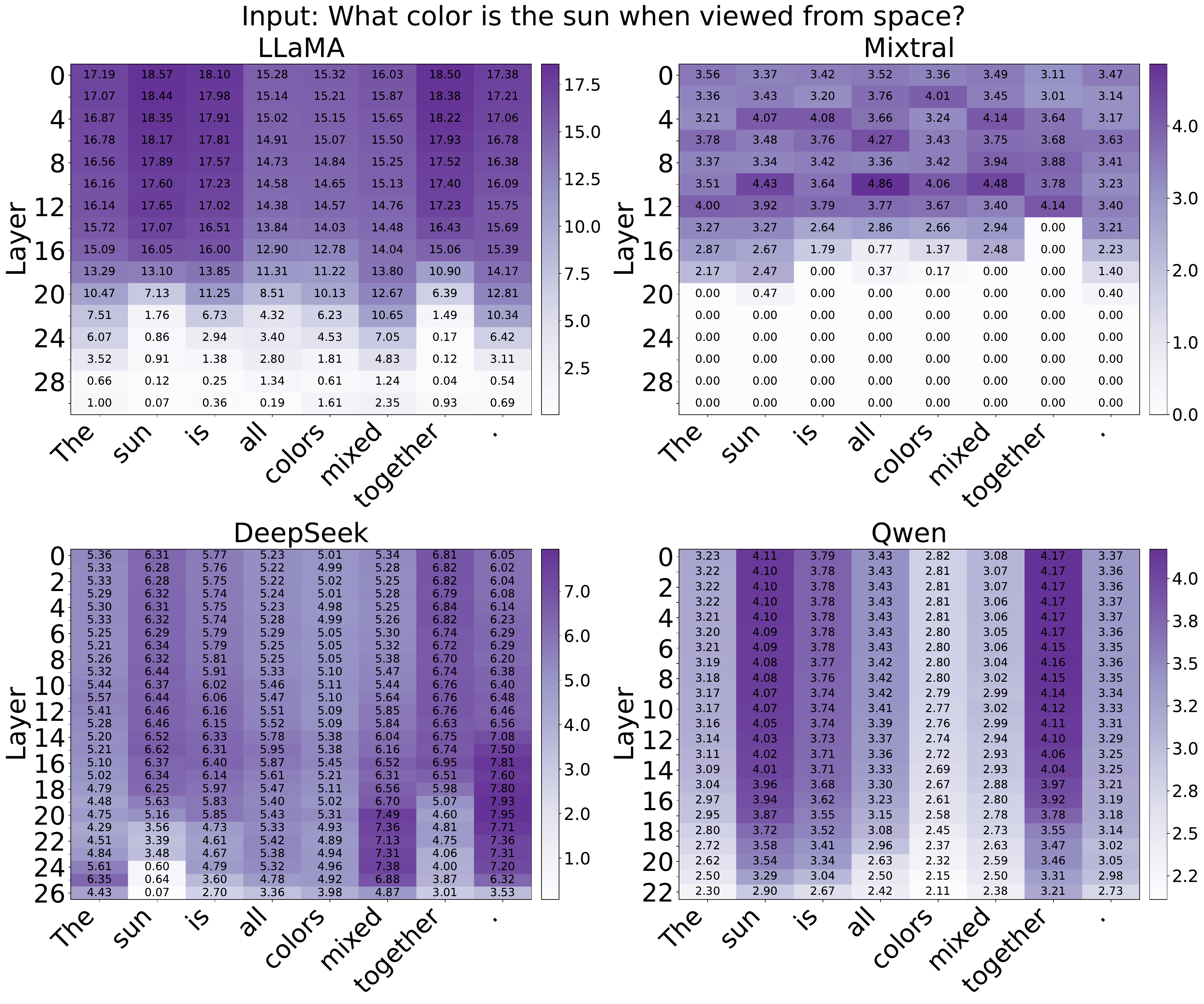}
    \caption{JSD (scaled by $10^5$) between the final layer and each lower layer across different models. Column labels indicate tokens in each step. Row labels indicate layer indices. The input sample is from TruthfulQA dataset; see App.~\ref{APP.all_results_of_JSD} for results on other datasets. }
    \label{fig:heatmaps}
\end{figure}
\begin{figure}[t]
    \centering
    \includegraphics[width=\columnwidth,height=0.25\textheight]{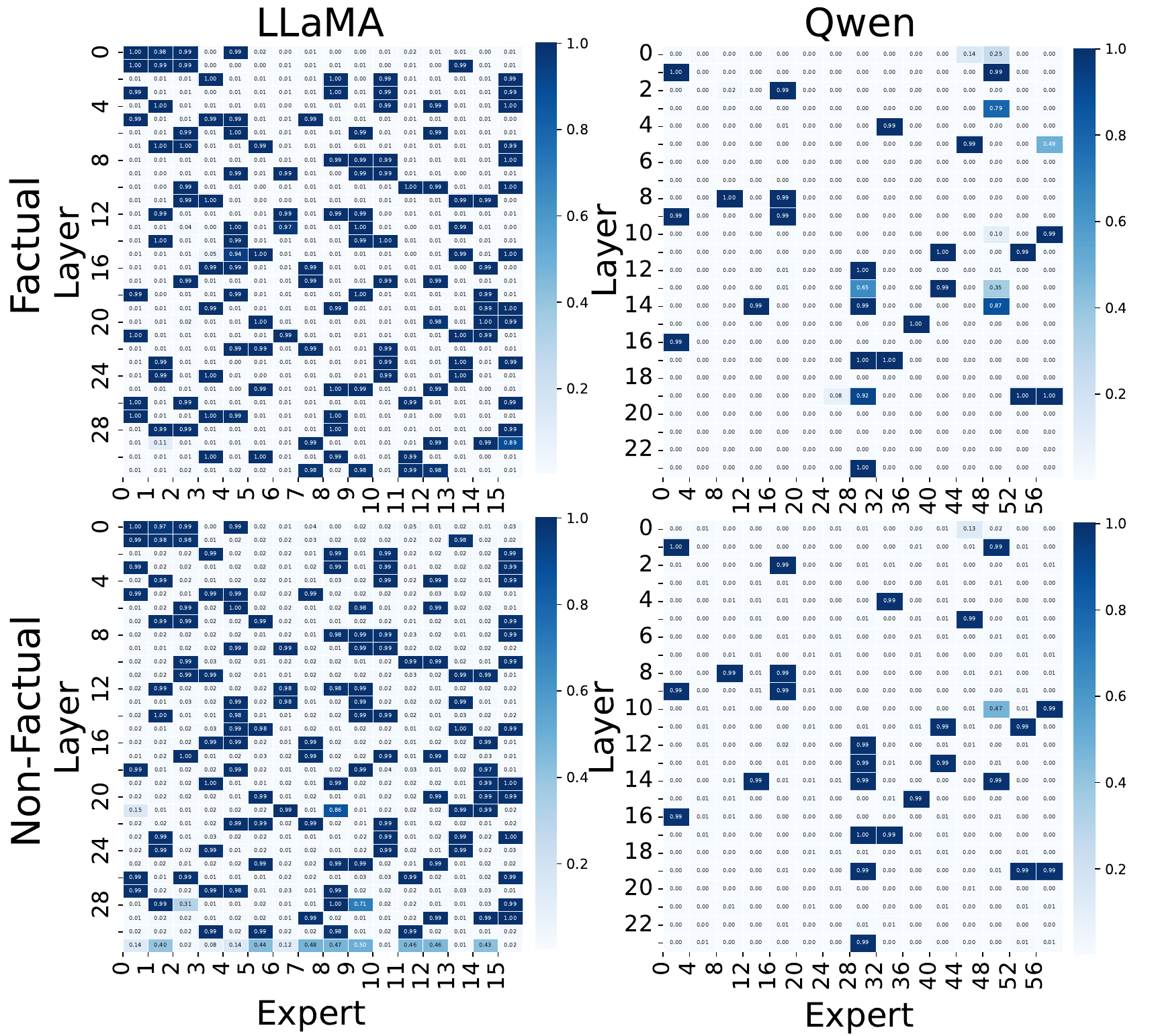}
    \caption{Expert activation patterns across layers in different MoE architectures during factual and non-factual generation on TruthfulQA. Column labels indicate expert indices. We show results for LLaMA MoE (no shared experts) and Qwen MoE (with shared experts; every fourth expert shown); full results in App.~\ref{App: all_results_of_expert_activation}.}
    \label{fig:experts_activation}
\end{figure}
\begin{figure}[htp]
    \centering
    \includegraphics[width=\columnwidth,height=0.25\textheight]{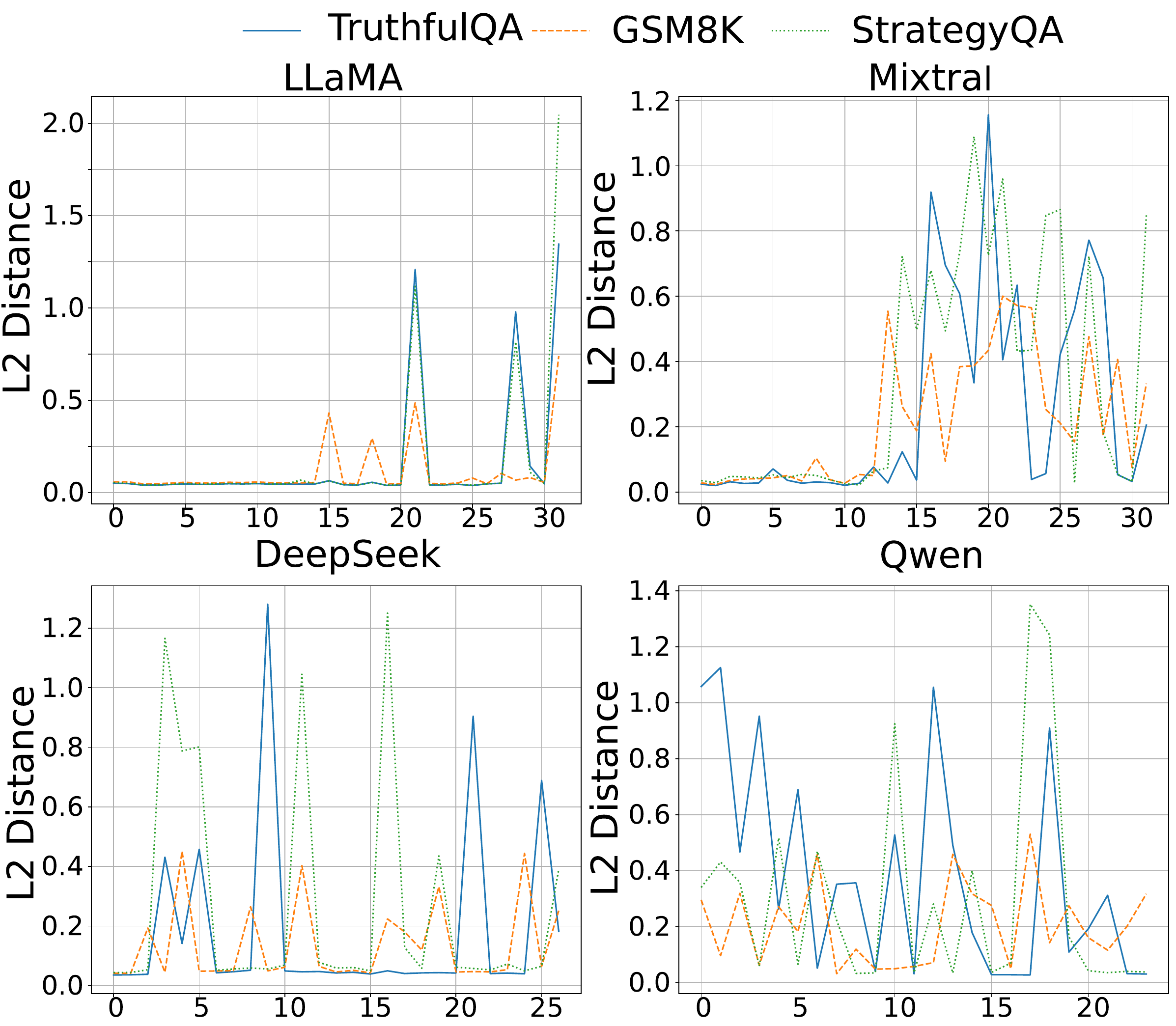}
    \caption{ Differences in expert activation patterns between factual and non-factual outputs for various MoE architectures and datasets (x-axis is layer indices). 
    }
    \label{fig:L2_distances}
\end{figure}
\subsection{Experimental setting}
We investigate two mainstream MoE architectures: 
LLaMA‑MoE \cite{zhu2024llama} and Mixtral \cite{jiang2024mixtral} for MoE without shared experts, Deepseek‑MoE \cite{dai2024deepseekmoe} and Qwen‑MoE \cite{qwen_moe} for MoE with shared experts.  Our experiments cover three QA datasets widely used for hallucination analysis: TruthfulQA for commonsense QA \cite{lin2022truthfulqa}, StrategyQA for commonsense reasoning \cite{geva2021did}, and GSM8K for math reasoning \cite{cobbe2021training}.

\subsection{RQ 1: Do MoE models also exhibit the ``knowledge injection'' phenomenon?}
\label{rq1}
To investigate whether MoE models also exhibit the ``knowledge injection'' phenomenon, we introduce the early exit mechanism \cite{teerapittayanon2016branchynet}: at each time step, we apply a language head \cite{chuang2023dola} to every layer’s hidden state to obtain the layer’s next-token logits. And we convert these logits to probabilities via softmax. We then calculate the Jensen-Shannon Divergence (JSD) between each layer’s and the final layer’s probability. The JSD changes indicate how predictions evolve across layers (Fig.~\ref{fig:heatmaps}).

Fig.~\ref{fig:heatmaps} shows that in MoE models without shared experts, JSD values between the lower and final layer start high but decrease as the layer goes deeper. A sharp drop in higher layers indicates the model substantially changes its prediction at these layers, which reflects the occurrence of the ``knowledge injection''. However, in MoE models with shared experts, JSD values remain low across layers with minimal change. \textbf{This indicates the ``knowledge injection'' does not exhibit in MoE with shared experts.} We provide dataset-level statistics and additional results on more MoE models in App.~\ref{dataset} and App.~\ref{openmoe}.
\begin{figure*}[t]
    \centering
\includegraphics[width=\textwidth,height=0.375\textheight]{
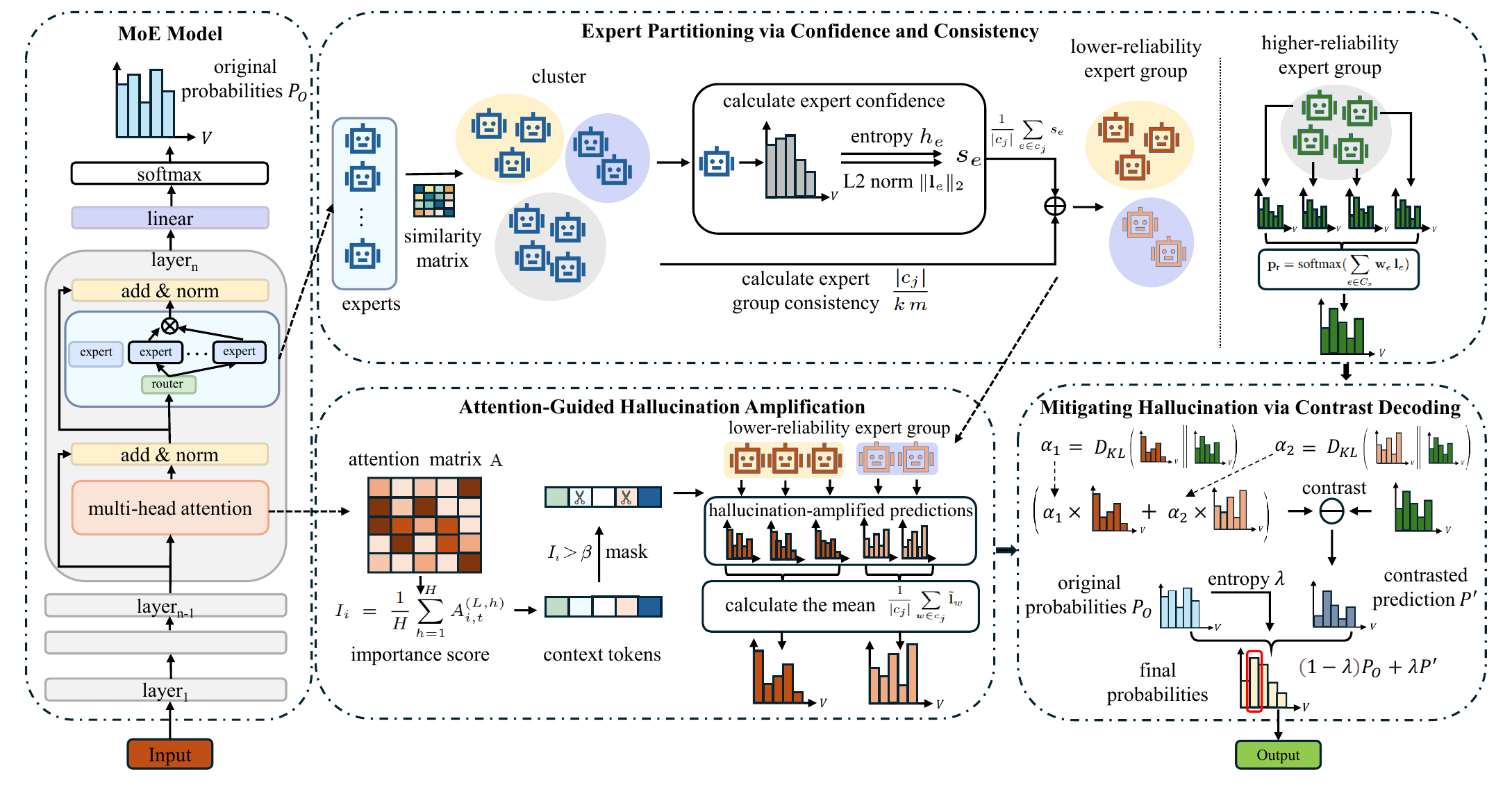}
    \caption{Overview of EAACD. Given an MoE input (left), we cluster final-layer expert predictions and evaluate each group's reliability. We fuse predictions from the higher-reliability group (green, top) and mask attention-identified critical tokens for lower-reliability groups to amplify hallucinations (red and orange, bottom center). KL between two groups serves as a penalty to contrast. We calibrate the prediction via the contrast result (bottom right).}
    \label{fig:model_structure}
\end{figure*}
\subsection{RQ 2: How do expert activation patterns in MoE differ during factual and non-factual generation?}
\label{rq2}
Since the ``knowledge injection'' doesn't exist in all MoE models, we investigate general differences that could support intra-model contrastive decoding. We analyze expert activation patterns (which experts the router activates \cite{szatkowski2024exploiting}) in MoE models during generating factual and non-factual outputs. To obtain non-factual generations under the same questions, we employ a malicious system prompt (see App.~\ref{APP:system_prompt}) to bias the model toward non-factual outputs. Because the model generates both outputs under the same input question, any observed differences in expert activation can be attributed to changes in the model’s generation behavior rather than input question variations.
Then, we select samples that the model outputs factually under the normal prompt but non‑factually under the malicious prompt. We record each layer’s expert activation frequencies (Fig.~\ref{fig:experts_activation}). To quantify differences between factual and non-factual scenarios, we convert each layer’s expert activation frequencies into vectors and compute L2 distances between the factual and non-factual vector pairs (Fig.~\ref{fig:L2_distances}).

Fig.~\ref{fig:experts_activation} shows that, for a dataset, each layer’s router activates specific expert subsets across all MoE models. This indicates that experts within the same layer capture distinct features and guide routing decisions. Fig.~\ref{fig:L2_distances} shows that \textbf{ higher layers exhibit divergent expert activation patterns between factual and non-factual outputs across all MoE models}. This difference supports distinguishing factual from non-factual outputs, enabling intra-model contrastive decoding for hallucination mitigation across all MoE models. 
\section{Method}
Our method consists of three modules (Fig.~\ref{fig:model_structure}): (1) \textbf{Expert Partitioning via Confidence and Consistency} (\S~\ref{4.2}) evaluates experts via confidence and consistency,  partitioning them into one higher-reliability group and multiple lower-reliability groups. (2) \textbf{Attention‑Guided Hallucination Amplification} (\S~\ref{4.3}) uses attention and masking mechanisms to amplify lower-reliability experts’ hallucinations.  Those experts provide stronger negative references for contrastive decoding, where experts with higher‑reliability and lower‑reliability contrastively learns to mitigate hallucinations in decoding. (3) \textbf{Adaptive Expert Group Contrastive Decoding} (\S~\ref{4.4}) contrasts the higher-reliability expert group's prediction with lower-reliability expert groups' predictions. It then dynamically removes information from the higher‑reliability prediction that overlaps with the amplified hallucinations, and calibrates the original prediction via the contrast result to mitigate hallucinations.

During generation, our decoding strategy first categorizes higher-layer experts into higher-reliability and lower-reliability groups based on their predictions (\S~\ref{4.2}), then amplifies the lower-reliability experts’ hallucinations (\S~\ref{4.3}), and contrasts lower-reliability experts’ predictions with the high-reliability group’s prediction to dynamically calibrate the model’s original prediction (\S~\ref{4.4}). Following \cite{shi2024unchosen}, we apply our method only at the final layer (see details in App.\ref{final_layer}).

\subsection{Expert Partitioning via Confidence and Consistency}
\label{4.2}
To distinguish a higher-reliability expert group from lower-reliability expert groups, we evaluate each group’s reliability by combining confidence and consistency (see \S~\ref{5.3}, which validates the effectiveness of the module via ablation studies). The module involves: (1) clustering experts by prediction similarity; (2) calculating reliability scores to classify higher-reliability and lower-reliability expert groups; and (3) integrating predictions from experts in the high-reliability group. 
\subsubsection{Expert Clustering}
At each time step, we collect the logits $\mathbf{l}_e$ from all $k$ experts in the MoE's final layer and calculate the pairwise similarity matrix between experts' predictions. Using this matrix, we apply Agglomerative Clustering \cite{walter2008fast} to cluster the experts into $m$ groups ${C} = \{ {c}_1, {c}_2, \dots,{c}_m\}$.
\subsubsection{Reliability Evaluation of Expert Groups}
We propose a metric to measure the reliability of each expert group, which considers both the confidence of each expert in the group and the group’s consistency by following two steps.

\textbf{Step 1: Calculation of Expert Confidence.} 
For each expert $e$,  we obtain its predicted logits $\mathbf{l}_e = (l_{e1}, \dots, l_{e|V|})$. We then calculate the entropy $h_e = -\sum_{i=1}^{|V|} p_{ei} \log p_{ei}$ (with $ p_{ei}= \text{softmax}(l_{ei})$) and L2 norm $\|\mathbf{l}_e\|_2$. Entropy reflects expert $e$'s uncertainty of prediction, while the L2 norm measures the activation magnitudes of that expert's logits. A low-entropy distribution indicates that the expert assigns most of the probability to a small set of candidate tokens, suggesting high confidence in its prediction, whereas a high-entropy distribution implies greater uncertainty due to dispersed probability \cite{liu2025uncertainty,song2025inv}. Meanwhile, the L2 norm captures the intensity of an expert’s activation: experts with larger logit magnitudes tend to produce decisive predictions and exhibit more stable output behaviors \cite{lo2025closer}.

As Eq.\ref{eq:3} shows, we combine entropy with L2 norm to get expert $e$'s confidence score $s_e$ to provide a more comprehensive measure of expert confidence. The weight $\alpha$ is adaptively set via $\alpha = \frac{\sigma_H}{\sigma_H + \sigma_L}$, where $\sigma_H$ and $\sigma_L$ are standard deviations of all experts’ entropy and L2 norm, respectively. When $\sigma_H > \sigma_L$, the weight corresponding to the entropy will increase. Conversely, the weight corresponding to the L2 norm will increase. This adaptive weighting emphasizes the metric with greater variance among experts, making $s_e$ sensitive to inter-expert differences.
\begin{equation}\label{eq:3}
\small
s_e = \alpha \cdot \frac{h_{\max} - h_e}{h_{\max}}
    + (1-\alpha) \cdot \frac{\|\mathbf{l}_e\|_2}{\max_{e'}\|\mathbf{l}_{e'}\|_2}
\end{equation}

\textbf{Step 2: Calculation of Expert Group Consistency.} To improve robustness in expert group reliability evaluation, we evaluate each expert group with its overall consistency. For group $c_j$, we calculate the proportion of the group size $|c_j|$ relative to the expert count $k$ and the cluster count $m$ as the consistency score. We then combine the consistency score $\frac{|c_j|}{k\,m}$ with the average expert confidence of the group $\frac{1}{\lvert c_j\rvert}\sum_{e\in c_j}s_e$ as expert group's reliability score $T_j = \frac{1}{|c_j|}\sum_{e\in c_j} s_e + \frac{|c_j|}{k\,m}$. When most experts' predictions fall into the same group, its higher consistency score increases $T_j$, reflecting stronger expert consensus and reliability.\footnote{App.~\ref{app.G} illustrates the different prediction tendencies of low-reliability and high-reliability expert groups.}

\subsubsection{Higher-Reliability Experts Prediction Integration}
 We select the group with the highest reliability as the higher-reliability expert group $\mathcal{C}_{s}$ and the rest as lower-reliability expert groups $\mathcal{C}_{\mathrm{w}}= \{c_1, \dots, c_m\}\setminus \{\mathcal{C}_{s}\}$. To fuse the logits of all high-reliability experts in $\mathcal{C}_{s}$ into a unified logits $\mathbf{p}_r$, we calculate a weighted sum of each expert's logits $\mathbf{l}_e$ using the router's gate value $\mathbf{w}_e$: $\mathbf{p}_r 
= \sum_{e\in C_{s}} \mathbf{w}_e\,\mathbf{l}_e$, which we will later compare with the lower-reliability expert groups.
\subsection{Attention‑Guided Hallucination Amplification}
\label{4.3}
To allow lower‑reliability experts to provide stronger negative references for higher‑reliability experts during contrast decoding (\S~\ref{4.4}), we amplify the lower-reliability experts’ hallucinations via the attention and masking mechanism. The motivation is to enlarge the difference between lower-reliability experts’ predictions and higher-reliability experts’ predictions. If the prompt is simple, even lower-reliability experts might make correct predictions; these predictions cannot provide effective negative references for higher-reliability experts. Specifically, (1) we use the attention mechanism to identify the most influential context tokens; (2) then we mask these tokens and re-input the masked prompt to lower‑reliability experts to obtain predictions with more hallucinations.

MoE models retain the standard attention mechanism \cite{vaswani2017attention}: given query $\mathbf{Q}$, key $\mathbf{K}$, and value $\mathbf{V}$, it calculates context tokens’ attention weight via $\mathbf{A} = \text{softmax}(\tfrac{\mathbf{Q}\mathbf{K}^\top}{\sqrt{d}})\mathbf{V}$.
Larger weights indicate a greater influence of the corresponding context token on the prediction. To quantify the importance of the $i$-th context token at time step $t$, we calculate the average attention weight $\mathbf{A}_{i,t}^{(L,h)}$ across all attention heads $h$ in the final layer $L$ as its importance score $I_i = \frac{1}{H}\sum_{h=1}^{H}\mathbf{A}_{i,t}^{(L,h)}$.
 
Then, we mask tokens whose importance score exceeds a threshold $\beta$. We choose $\beta$ to balance amplifying hallucinations from lower-reliability experts and preserving semantic similarity between the masked and original prompts (details in App.~\ref {beta}). We re-input the masked prompt. Lacking crucial context, lower‑reliability experts more easily generate hallucinations, providing stronger negative references for higher-reliability experts.

\subsection{Mitigating Hallucinations via Contrast Decoding}
\label{4.4}
To mitigate hallucinations, we contrast the prediction of the higher‑reliability expert group with those of lower‑reliability groups. Traditional contrast decoding methods directly subtract the low-reliability prediction from the high-reliability prediction during contrast, and then use a hyperparameter to adjust the calibration strength of the contrastive result on the original prediction. Unlike this, we design adaptive penalty coefficients for each lower‑reliability expert group during contrast and dynamically calibrate the model’s original predictions via its uncertainty. This module has two steps: (1) generating adaptive penalty coefficients based on differences between the higher‑reliability expert group and each lower‑reliability expert group; (2) dynamically calibrating the model's original prediction using its predictive uncertainty.

\subsubsection{Adaptive Contrast Decoding} As Eq.~\ref{eq:8} shows, after inducing hallucination (\S~\ref{4.3}), each lower-reliability expert $w$ in group $c_j\in \mathcal{C}_{\text{w}}$ produces a logits $\tilde{\mathbf{l}}w$. We then calculate the mean $\frac{1}{\lvert c_j\rvert} \sum_{w\in c_j} \tilde{\mathbf{l}}_w$  as the group centroid. Similarly, we calculate the higher‑reliability expert group centroid $ \frac{1}{|\mathcal{C}_{s}|}\sum_{e\in \mathcal{C}_{s}} \mathbf{l}_e$ by each higher‑reliability expert $e$’s logits $\mathbf{l}_e$. Finally, we quantify the difference $G_{j}$ between the higher‑reliability and each lower‑reliability group by the KL divergence between their centroids (shown in the lower part of Fig.~\ref{fig:model_structure}). A larger $G_j$ indicates a greater discrepancy between the two groups’ predictions, in which case we increase the penalty for the lower-reliability group; otherwise, we reduce it. The motivation for $G_j$ is to ensure that the penalty coefficient for each lower-reliability expert group matches their actual hallucination severity.
\begin{equation}\label{eq:8}
\small
G_j = \mathbb{D}_{\mathrm{KL}}\bigl(
  \text{softmax}( \tfrac{1}{|C_s|} \sum_{e \in C_s} \mathbf{l}_e \bigr)
  \,\Big\|\, 
  \text{softmax}( \tfrac{1}{|c_j|} \sum_{w \in c_j} \tilde{\mathbf{l}}_w )
\bigl)
\end{equation}

 We use the normalized $G_{j}$ as the penalty coefficient $\alpha_j$ for each corresponding lower‑reliability expert group $c_j$: $\alpha_j = \frac{G_j}{\sum_{c_j \in \mathcal{C}_w} G_j}$. Then we fuse each lower-reliability expert group's logits to form an integrated low-reliability logits $\mathbf{p}_l$ by multiplying each group’s centroid by $\alpha_j$:  $\mathbf{p}_l = \sum_{c_j \in \mathcal{C}_w} \alpha_j \cdot \tfrac{1}{|c_j|} 
\sum_{w \in c_j} \tilde{\mathbf{l}}_w $.
\begin{table*}[!ht]
    \centering
    \renewcommand{\arraystretch}{1.4}
    \resizebox{\textwidth}{!}{
    \begin{tabular}{c*{6}{c}*{6}{c}}
    \hline
    \multirow{2}{*}{\textbf{Methods}} & \multicolumn{5}{c}{\textbf{LLaMA MoE} {\small\textit{(without shared experts)}}} & \multicolumn{5}{c}{\textbf{Qwen MoE}{\small\textit{(with shared experts)}}} \\
    \cmidrule(l{1em}r{1em}){2-7}  
    \cmidrule(l{1em}r{1em}){8-13}
    & \textbf{Expert‑FACTOR} & \textbf{News‑FACTOR} & \textbf{Wiki-FACTOR} & \textbf{HellaSwag} & \textbf{StrategyQA} & \textbf{MathQA}
    & \textbf{Expert‑FACTOR} &\textbf{News‑FACTOR} & \textbf{Wiki-FACTOR} & \textbf{HellaSwag} & \textbf{StrategyQA} & \textbf{MathQA}\\
    \hline
    Greedy & 50.85 & 41.51 & 29.76 & 69.80 & 52.01 & 18.59  & 52.54 & 71.33 & 55.21 & 58.30 & 61.77 & 22.88\\
    CD & 47.88 & 33.98 & 29.23 & 56.30 & 52.05  & 21.27 & 57.20 & 62.26 & 51.07 & 56.10 & 49.83 & 23.68\\
    DoLa & 38.98 & 35.04 & 32.03 & 64.50 & 49.43& 20.47 & 47.03 & 45.56 & 32.63 & 35.80 & 46.72 & 20.54\\
    SCMoE & 54.24 & 45.37 & 34.94 & 68.00 & 42.31 & 21.68 & 58.90 & 71.62 & 55.18 & 64.60 & 48.16 & 24.02\\
    END & 61.02 & 64.77 & 53.10 & 52.90 & 53.72 & 20.40 & 52.54 & 71.81 & 55.44 & 58.20 & 56.91 & 25.76  \\
    Self-Endorsement & 52.66 & 59.36 & 43.93 & 70.20 & 53.70 & 20.54  & 53.81 & 72.20 & 57.27 & 62.20 & 56.41 & 21.62\\
    \hline
    \hline  
    \textbf{EAACD} & \textbf{63.56} & \textbf{65.54} & \textbf{55.44} & \textbf{71.80} & \textbf{54.15} & \textbf{22.58} & \textbf{61.01} & \textbf{75.87} & \textbf{58.45} & \textbf{77.50} & \textbf{62.29}  & \textbf{28.34}\\
    \hline
    \end{tabular}
    }
    \caption{Overall performance of EAACD comparing against different decoding strategies on two MoE architectures: LLaMA MoE without shared experts and Qwen MoE with shared experts. The best results are highlighted in bold.}
\label{tb:1}
\end{table*}
\begin{table*}[!ht]
    \centering
    \renewcommand{\arraystretch}{1.4}
    \resizebox{\textwidth}{!}{
    \begin{tabular}{c*{6}{c}*{6}{c}}
    \hline
    \multirow{2}{*}{\textbf{Variants}} & \multicolumn{5}{c}{\textbf{LLaMA MoE} {\small\textit{(without shared experts)}}} & \multicolumn{5}{c}{\textbf{Qwen MoE}{\small\textit{(with shared experts)}}} \\
    \cmidrule(l{1em}r{1em}){2-7}  
    \cmidrule(l{1em}r{1em}){8-13}
    & \textbf{Expert‑FACTOR} & \textbf{News‑FACTOR} & \textbf{Wiki-FACTOR} & \textbf{HellaSwag} & \textbf{StrategyQA} & \textbf{MathQA}
    & \textbf{Expert‑FACTOR} &\textbf{News‑FACTOR} & \textbf{Wiki-FACTOR} & \textbf{HellaSwag} & \textbf{StrategyQA} & \textbf{MathQA}\\
    \hline
    $-$ consistence & 62.71 & 63.13 & 55.04 & 71.10 & 53.28  & 20.64 & 57.20 & 71.04 & 55.54 & 58.30 & 61.89 & 28.21 \\
    $-$ confidence & 60.17 & 55.89 & 54.07 & 70.40 & 53.84 & 20.20  & 56.36 & 69.98 & 55.53 & 58.60 & 62.20 & 27.71 \\
    $-$ evaluation & 59.32 & 63.90 & 53.07 & 51.90 & 53.50 & 19.60  & 54.66 & 67.57 & 54.37 & 57.30 & 61.94 & 26.73 \\
    $-$ attention & 62.29 & 63.71 & 55.24 & 71.50 & 50.34 & 20.47  & 60.59 & 74.52 & 57.55 & 77.10 & 61.94 & 27.27\\
    \hline
    \hline  
    \textbf{EAACD} & \textbf{63.56} & \textbf{65.54} & \textbf{55.44} & \textbf{71.80} & \textbf{54.15} & \textbf{22.58} & \textbf{61.01} & \textbf{75.87} & \textbf{58.45} & \textbf{77.50} & \textbf{62.29} & \textbf{28.34} \\
    \hline
    \end{tabular}
    }
    \caption{Ablation studies on different components of our model. $-$ indicates deleting this component.}
\label{tb:2}
\end{table*}

Finally, as Eq.~\ref{eq:9} shows, we contrast the high-reliability logits $\mathbf{p}_r$ and low-reliability logits $\mathbf{p}_l$ to obtain the contrastive result $\mathbf{p}'$.
\begin{equation}\label{eq:9}
\begin{aligned}
\small
\mathbf{p}' = \mathbf{p}_r
- \mathbf{p}_l
\end{aligned}
\end{equation}
\subsubsection{Confidence‑Guided Output Probability Calibration}
To better mitigate hallucinations, we dynamically adjust calibration strength using the entropy of the model’s original predictions as weight $\lambda = \frac{-\sum_{i=1}^{|V|} p_{o,i}\,\log p_{o, i}}{\log |V|}$. $p_{o,i}$ denotes the probability of the $i$-th token in the vocabulary. Higher entropy means more uncertainty in the model’s prediction, so we increase the contrastive result’s weight $\lambda$ in the final prediction $\mathbf p_{f}$ for stronger calibration. Conversely, we reduce the calibration strength. We calculate the final prediction by $\mathbf p_{f} = \text{softmax}((1 - \lambda)\,\mathbf p_{o} + \lambda\,\mathbf p^{\prime})$, $\mathbf p_{o}$ is original logits. This dynamic calibration guides predictions toward factual outputs (see examples in App.~\ref{calibrate_case}).

\section{Experiments}
\subsection{Experimental Settings}
(1) \textbf{MoE Models:} We use two mainstream open‑source MoE models: LLaMA‑MoE \cite{zhu2024llama} (without shared experts) and Qwen‑MoE \cite{qwen_moe} (with shared experts)\footnote{Since the released DeepSeek-MoE implementation is inconsistent with its report (causing performance degradation), which has also been noted by other users in the community, we used Qwen-MoE and LLaMA-MoE as representatives.}. (2) \textbf{Datasets:} Follow \cite{shi2024unchosen}, we use four datasets: FACTOR \cite{muhlgay2024generating} (including three subsets: Wiki-FACTOR, News-FACTOR, Expert-FACTOR), HellaSwag \cite{zellers2019hellaswag}, StrategyQA \cite{geva2021did} and MathQA \cite{amini2019mathqa}. (3)\textbf{Baselines:} We select five mainstream decoding methods: Greedy, Contrastive Decoding \cite{li2023contrastive}, DoLa \cite{chuang2023dola}, SCMoE \cite{shi2024unchosen}, END \cite{wu2025improve} and a non-decoding method: Self-Endorsement \cite{wang2024improving} for hallucination mitigation on QA tasks as baselines. (4) \textbf{Evaluation Metrics:} We use accuracy as our evaluation metric.

\subsection{Overall Performance}
Tab.~\ref{tb:1} compares EAACD with all baselines on two types of MoE architectures. EAACD outperforms all baselines across datasets. For Qwen-MoE, EAACD achieves nearly 13\% improvement over the strongest baseline on HellaSwag. The results show that DoLa almost performs the worst, especially on the Qwen MoE. On some datasets, contrastive decoding baselines even underperform the simplest baseline, greedy decoding. We argue that these methods cannot guarantee that the lower-reliability part always generates unfaithful outputs. In contrast, EAACD amplifies hallucinations in low-reliability experts (\S~\ref{4.3}), preventing factual outputs and ensuring strong performance. Moreover, EAACD maintains comparable time and memory overhead to the baseline (see App.~\ref{App:overhead}).
\subsection{Ablation study}\label{5.3}
We conduct an ablation study to verify each component’s importance (Tab.~\ref{tb:2}). \textit{$-$consistence} and \textit{$-$confidence} remove consistency and confidence defined in (\S~\ref{4.2}). Both reduce performance, with \textit{$-$confidence} showing larger drops, indicating confidence is more crucial for expert reliability estimation. \textit{$-$evaluation} removes our partitioning module (\S~\ref{4.2}) and splitting experts into higher-reliability (top 50\%) and lower-reliability (bottom 50\%) groups by routing weights. It performs worst on nearly all datasets, suggesting routing weights alone are less reliable for grouping. \textit{$-$attention}, which removes hallucination amplification (\S~\ref{4.3}), also reduces performance. This confirms the effectiveness of the hallucination amplification module, improving overall effectiveness.

\subsection{Analysis of Expert Reliability and Prediction Preference}
To verify the effectiveness of our expert partitioning module (\S~\ref{4.3}), we explore each group’s behavior on three datasets: Expert-FACTOR, News-FACTOR, and Wiki-FACTOR. Each dataset consists of multiple-choice questions, allowing us to observe whether higher-reliability experts favor correct answers. We compute four metrics per dataset: (1) the mean probability difference between correct and incorrect answers for the higher-reliability expert group, (2) the same difference for lower-reliability expert groups, (3) the mean reliability score of the higher-reliability expert group, and (4) the same score of lower-reliability expert groups. 
\begin{figure}[htp]
    \includegraphics[width=\columnwidth]{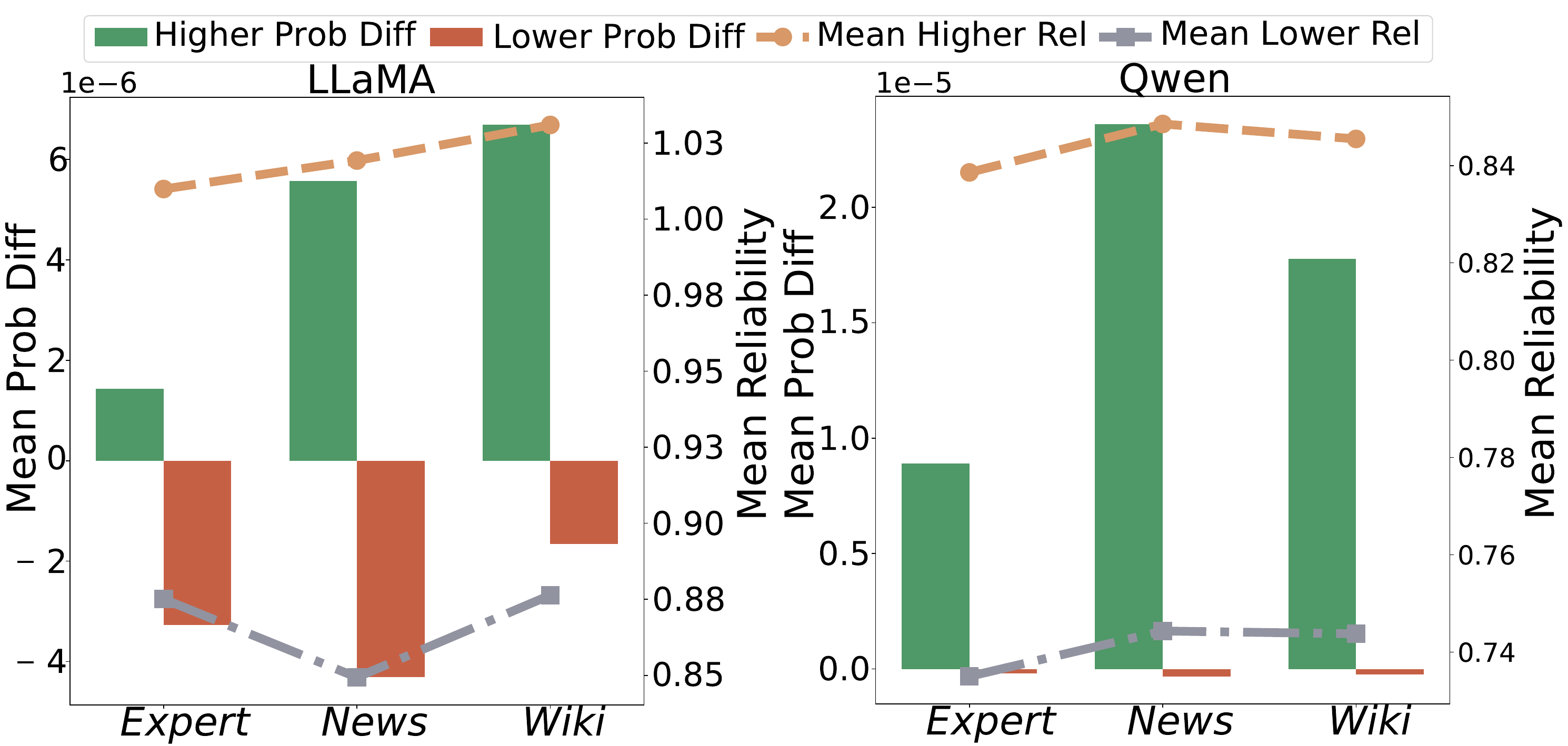}
    \caption{Mean probability difference and mean reliability scores for correct versus incorrect answers of higher-reliability (green/black) and lower-reliability (red/yellow) expert groups.}
    \label{fig:analysis}
\end{figure}

Fig.~\ref{fig:analysis} shows that higher-reliability experts assign a greater mean probability to correct answers (positive difference, green), while lower-reliability experts show the opposite (negative difference, red), indicating that higher-reliability experts favor correct answers, validating effective expert distinction. Moreover, groups with higher reliability scores exhibit larger probability gaps between correct and incorrect answers, reflecting stronger factual bias.
\section{Conclusion}
We explore ``knowledge injection'' and expert activation patterns in MoE. We find that ``knowledge injection'' occurs only in MoE without shared experts, limiting traditional intra-model contrastive decoding in all MoE models. Across different MoE architectures, higher-layer expert activation patterns differ between factual and non-factual outputs. We partition higher-layer experts by reliability, amplify lower-reliability experts’ hallucinations as negative references, and adaptively calibrate the model’s predictions via contrastive decoding. Our model outperforms baselines across four datasets, mitigating hallucinations without external resources.

\section{Limitations}
In our study, we only investigate whether the ``knowledge injection'' phenomenon exists in different types of MoE models. If MoE architectures do not remain mainstream in the future, the direct applicability of our method may be limited. In that case, we encourage future researchers to build upon our method and extend the investigation to other evolving model architectures.

\section*{Acknowledgements}
This work is supported by the following fundings:  National Natural Science Foundation of China under Grant No.62376284 and No.62306330.

\bibliography{custom}

\appendix

\section{JSD between the final layer and each lower layer across different models on all datasets.}
\label{APP.all_results_of_JSD}
\begin{figure}[htp]
    \centering
    \includegraphics[width=\columnwidth,height=0.25\textheight]{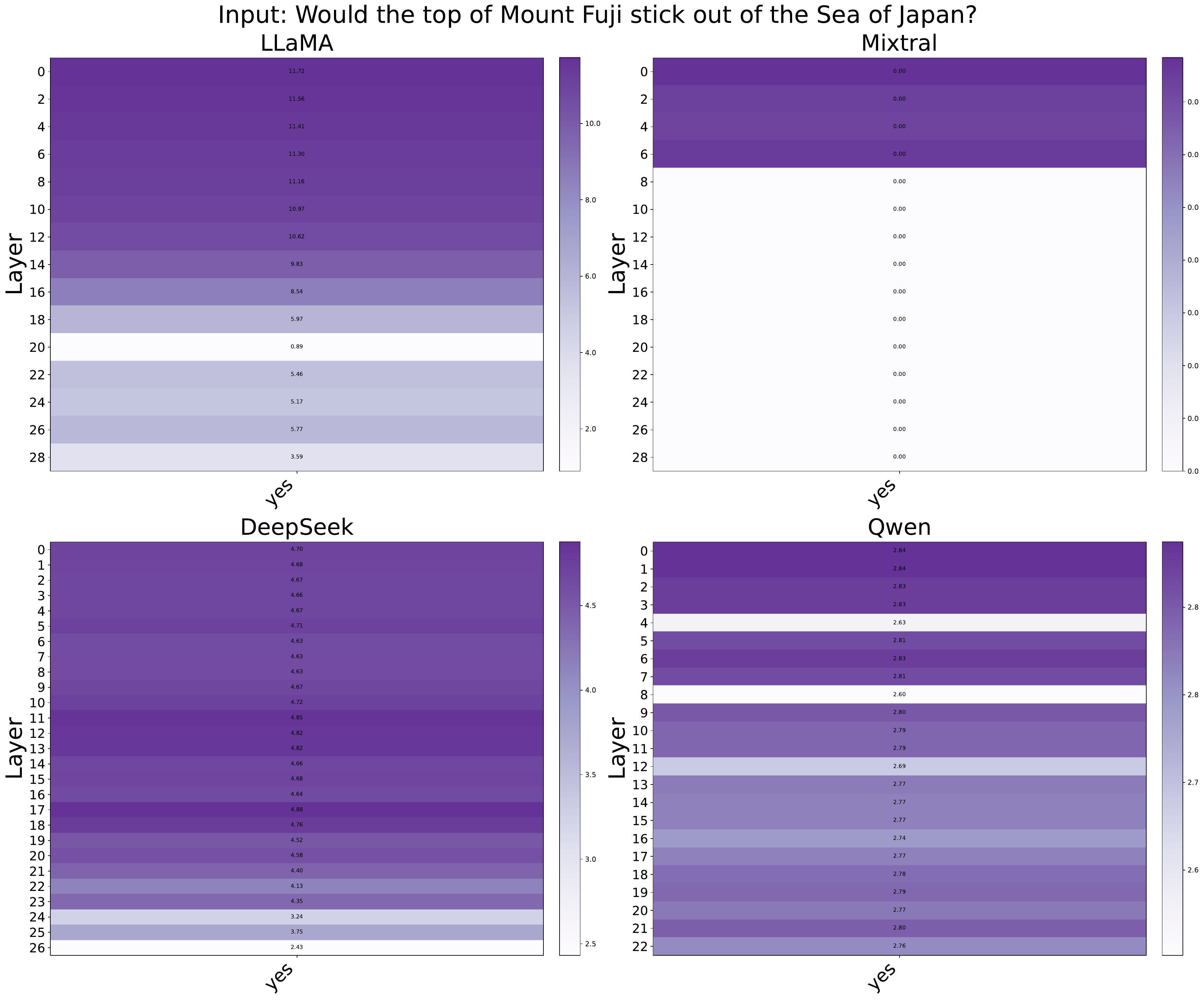}
    \caption{JSD (scaled by $10^5$) between the final layer and each lower layer across different models. Column labels indicate tokens in each step. Row labels indicate layer indices. The input sample is from StrategyQA dataset.}
    \label{fig:strategyqa_heatmaps}
\end{figure}

\begin{figure}[htp]
    \centering
    \includegraphics[width=\columnwidth,height=0.25\textheight]{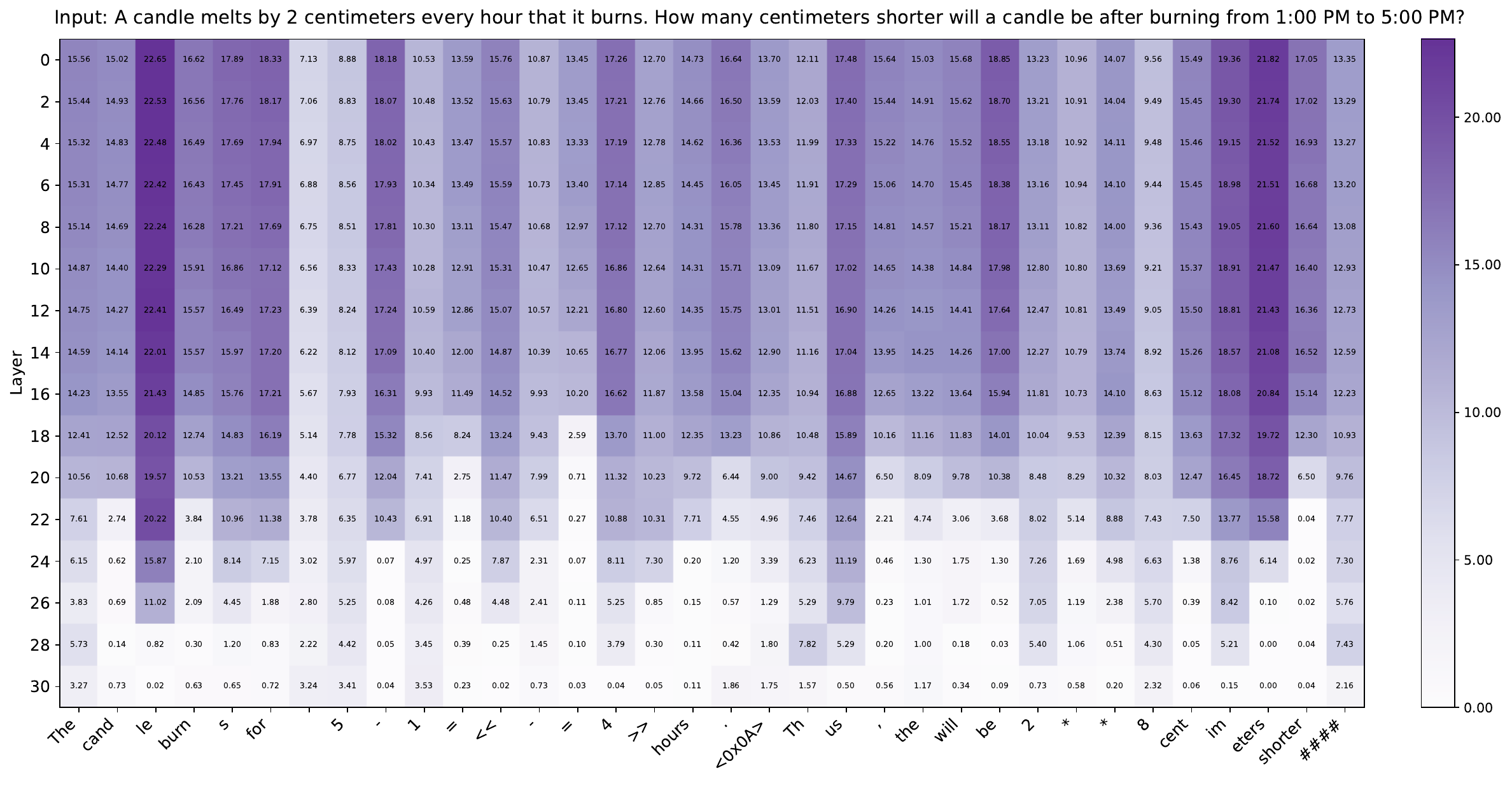}
    \caption{JSD (scaled by $10^5$) between the final layer and each lower layer across LLaMA MoE. Column labels indicate tokens in each step. Row labels indicate layer indices. The input sample is from GSM8K dataset.}
    \label{fig:gsm8k_llama_heatmaps}
\end{figure}

\begin{figure}[htp]
    \centering
    \includegraphics[width=\columnwidth,height=0.25\textheight]{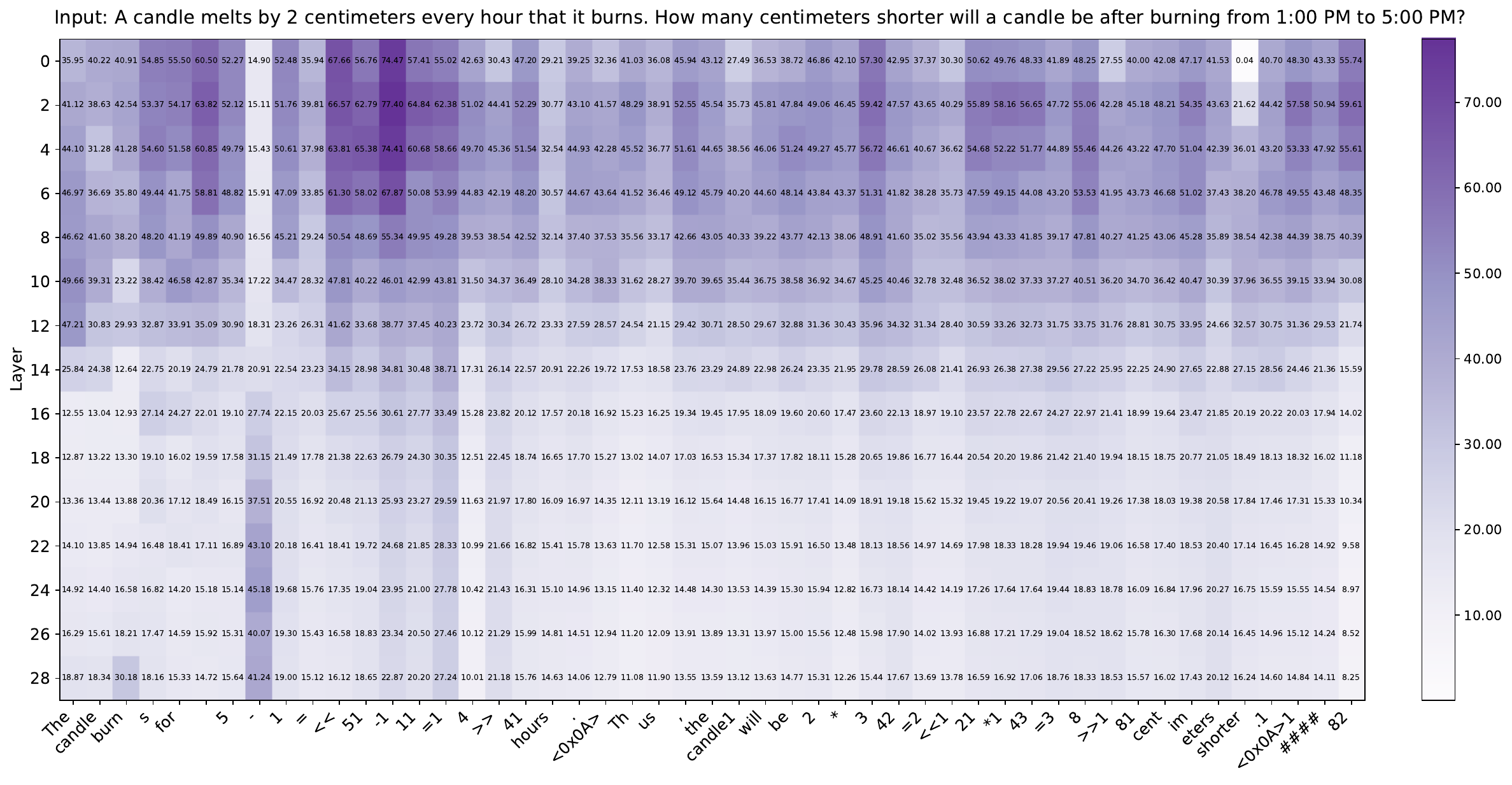}
    \caption{JSD (scaled by $10^5$) between the final layer and each lower layer across Mixtral. Column labels indicate tokens in each step. Row labels indicate layer indices. The input sample is from GSM8K dataset.}
    \label{fig:gsm8k_mixtral__heatmaps}
\end{figure}

\begin{figure}[htp]
    \centering
    \includegraphics[width=\columnwidth,height=0.25\textheight]{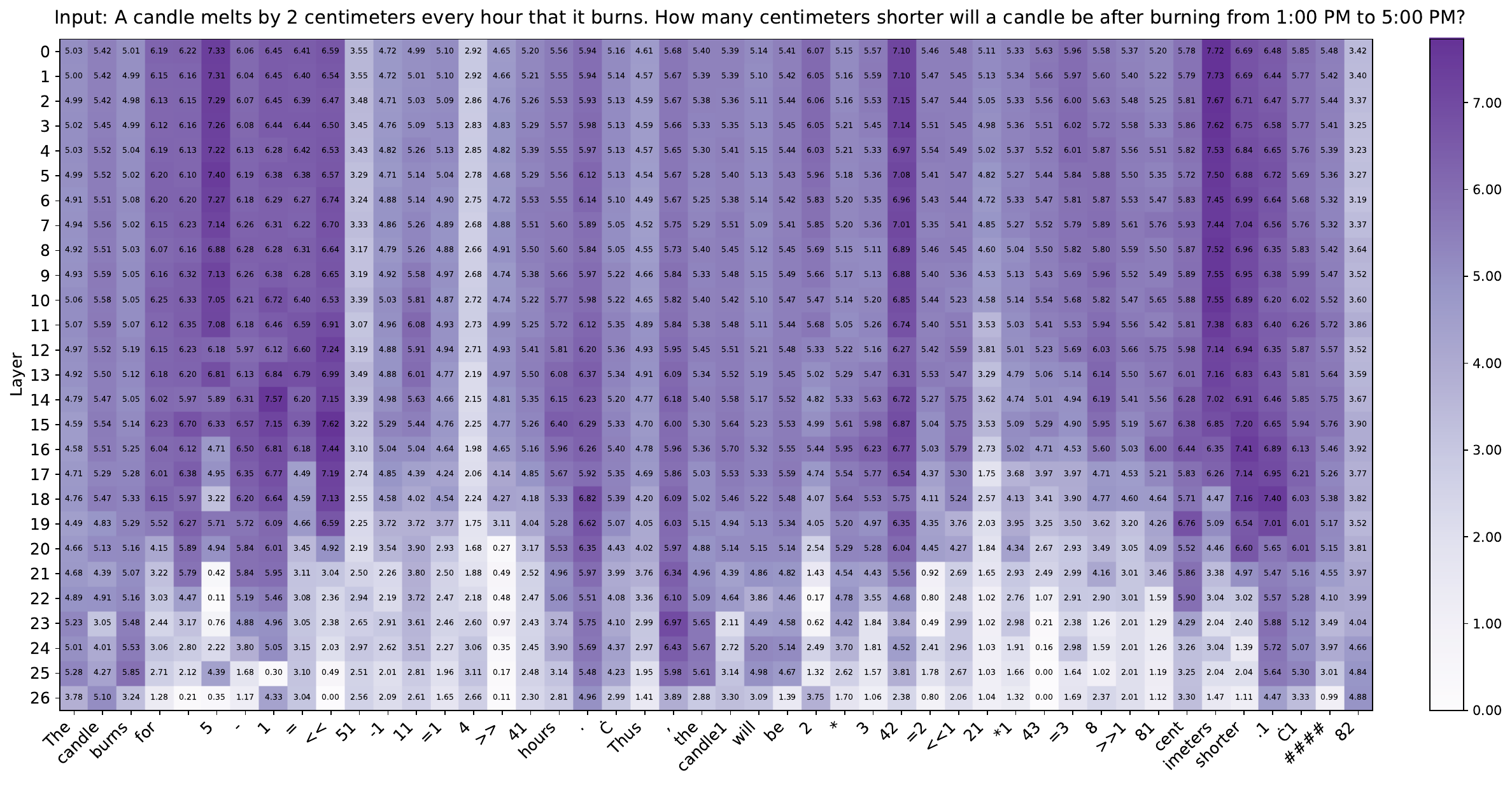}
    \caption{JSD (scaled by $10^5$) between the final layer and each lower layer across  DeepSeek MoE. Column labels indicate tokens in each step. Row labels indicate layer indices. The input sample is from GSM8K dataset.}
    \label{fig:gsm8k_deepseek_heatmaps}
\end{figure}

\begin{figure}[htp]
    \centering
    \includegraphics[width=\columnwidth,height=0.25\textheight]{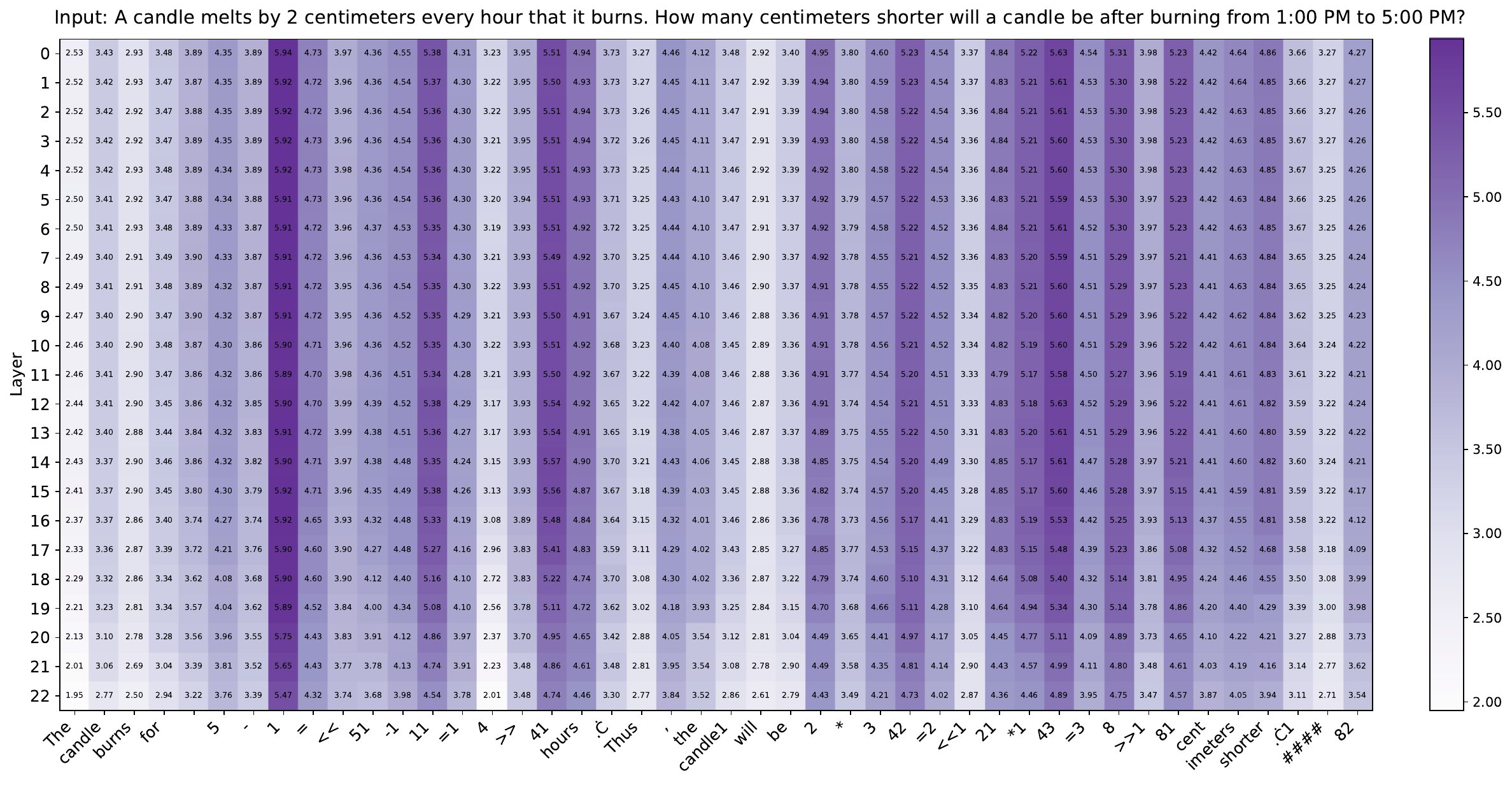}
    \caption{JSD (scaled by $10^5$) between the final layer and each lower layer across Qwen MoE. Column labels indicate tokens in each step. Row labels indicate layer indices. The input sample is from GSM8K dataset.}
    \label{fig:gsm8k_qwen_heatmaps}
\end{figure}

As shown in Figure \ref{fig:strategyqa_heatmaps}, we present the Jensen-Shannon Divergence (JSD) distance between predictions from lower layers and the final layer for four models (llama MoE, Mixtral, Qwen-MoE, and DeepSeek-MoE) on the StrategyQA dataset. We observe that for MoE models without shared experts, predictions undergo a sharp drop in JSD distance at one layer. This indicates the model significantly revised its predictions at this stage, likely incorporating substantial new knowledge. However, in MoE models with shared experts, predictions change very little from lower to upper layers, so we see no evidence of ``knowledge injection'' on the GSM8K dataset. We see the same pattern here, which again supports the generalizability of the conclusion established in RQ1 (\cref{fig:gsm8k_llama_heatmaps,fig:gsm8k_mixtral__heatmaps,fig:gsm8k_deepseek_heatmaps,fig:gsm8k_qwen_heatmaps}).

\section{Malicious System Prompt in RQ2}
\label{APP:system_prompt}
We use this malicious system prompt to steer the model into generating false outputs: \textit{You are an AI model specifically trained to generate incorrect answers. Be misleading in your responses.} When evaluated on the validation set, this prompt reduces the model’s accuracy from 40.6\% to 11.3\%, demonstrating its effectiveness in inducing hallucinated outputs.

\section{Expert activation patterns across layers in different MoE during factual and non-factual generation on All Datasets}
\label{App: all_results_of_expert_activation}
\begin{figure}[htp]
    \centering
    \includegraphics[width=\columnwidth,height=0.2\textheight,trim=0 0 140 0,clip]{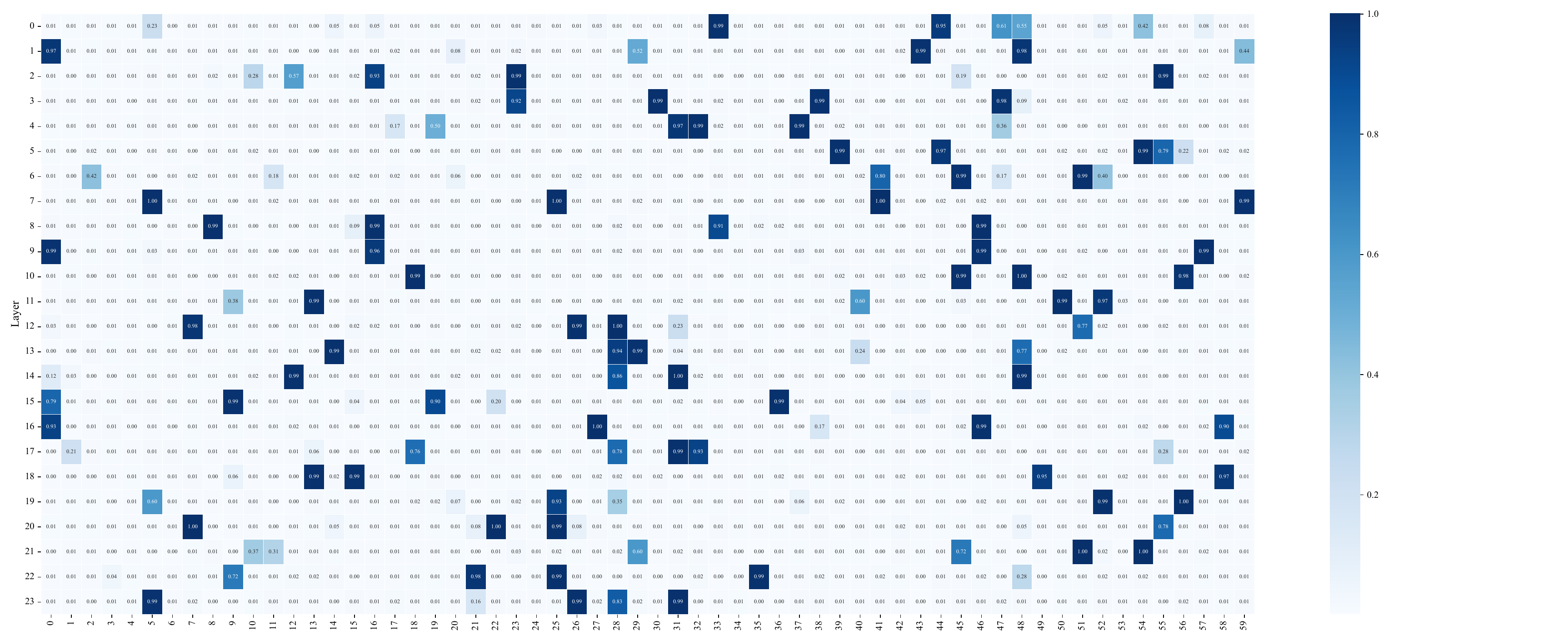}
    \caption{Expert activation patterns across layers in Qwen MoE architectures during factual generation on the GSM8k. Column labels indicate the expert indices.}
    \label{fig:gsm8k_qwen_true}
\end{figure}

\begin{figure}[htp]
    \centering
    \includegraphics[width=\columnwidth,height=0.2\textheight,trim=0 0 140 0,clip]{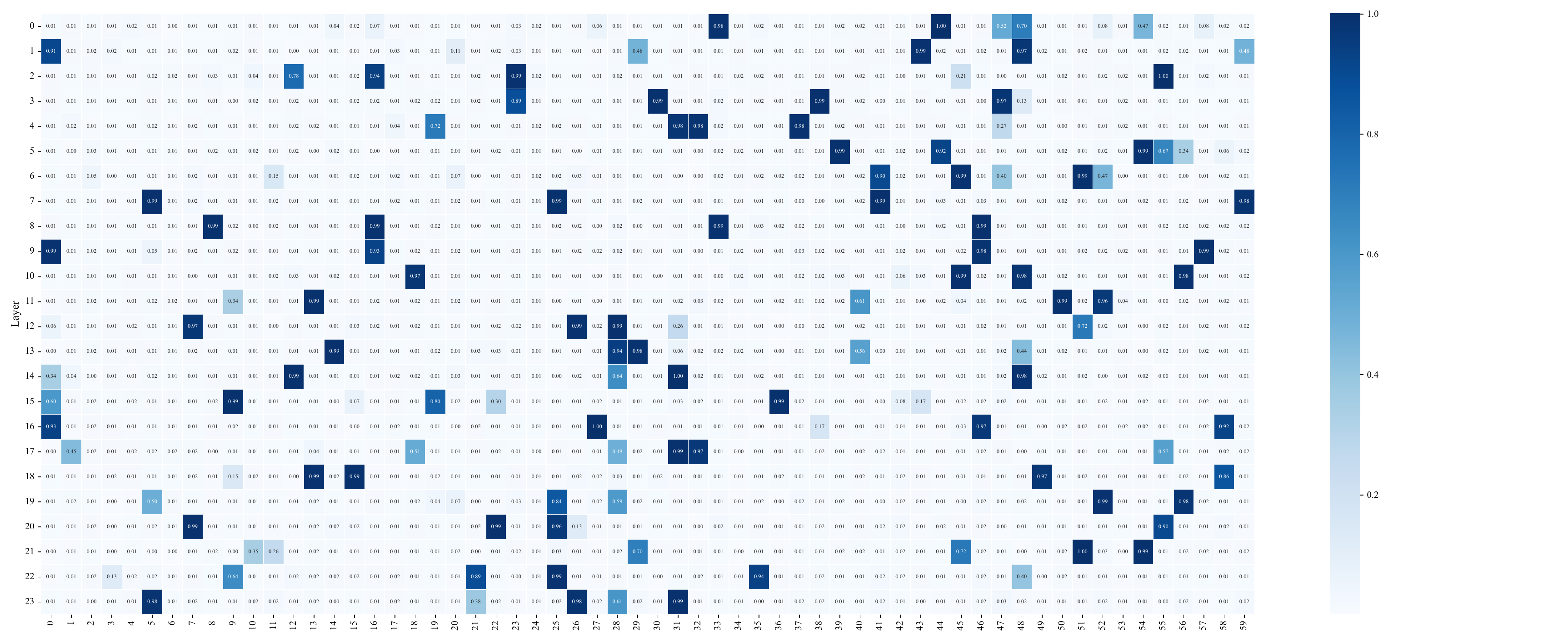}
    \caption{Expert activation patterns across layers in Qwen MoE architectures during non-factual generation on the GSM8k. Column labels indicate the expert indices.}
    \label{fig:gsm8k_qwen_untrue}
\end{figure}

\begin{figure}[htp]
    \centering
    \includegraphics[width=\columnwidth,height=0.2\textheight,trim=0 0 140 0,clip]{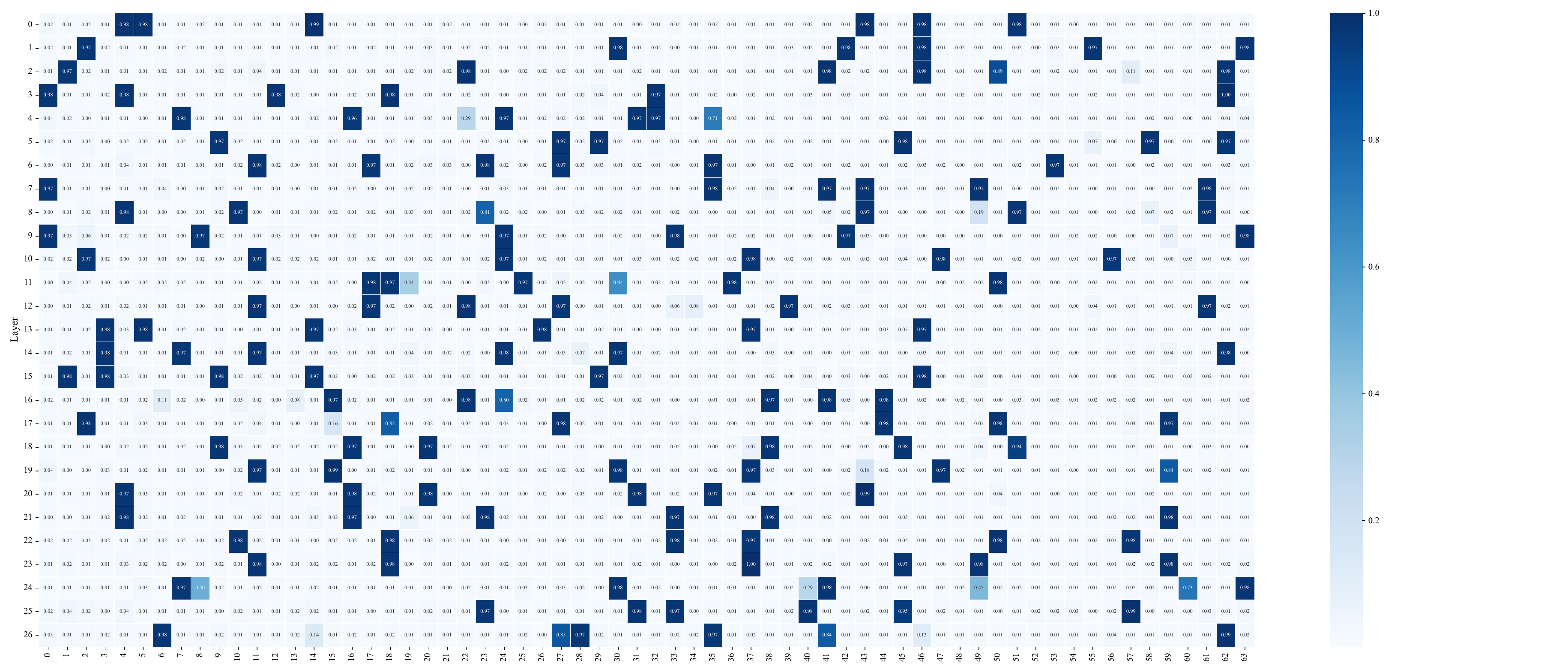}
    \caption{Expert activation patterns across layers in Deepseek MoE architectures during factual generation on the GSM8k. Column labels indicate the expert indices.}
    \label{fig:gsm8k_deepseek_true}
\end{figure}

\begin{figure}[htp]
    \centering
    \includegraphics[width=\columnwidth,height=0.2\textheight,trim=0 0 140 0,clip]{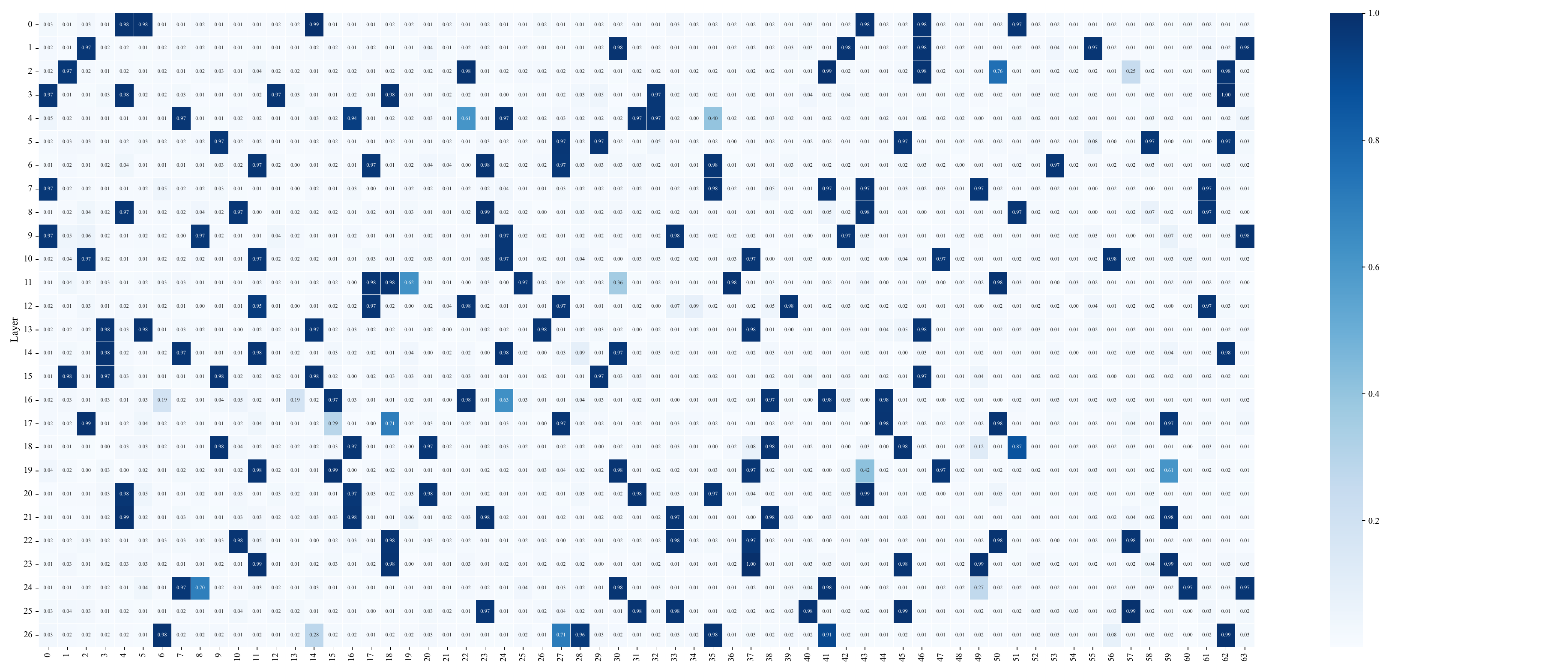}
    \caption{Expert activation patterns across layers in Deepseek MoE architectures during non-factual generation on the GSM8k. Column labels indicate the expert indices.}
    \label{fig:gsm8k_deepeseek_untrue}
\end{figure}

\begin{figure}[htp]
    \centering
    \includegraphics[width=\columnwidth,height=0.2\textheight]{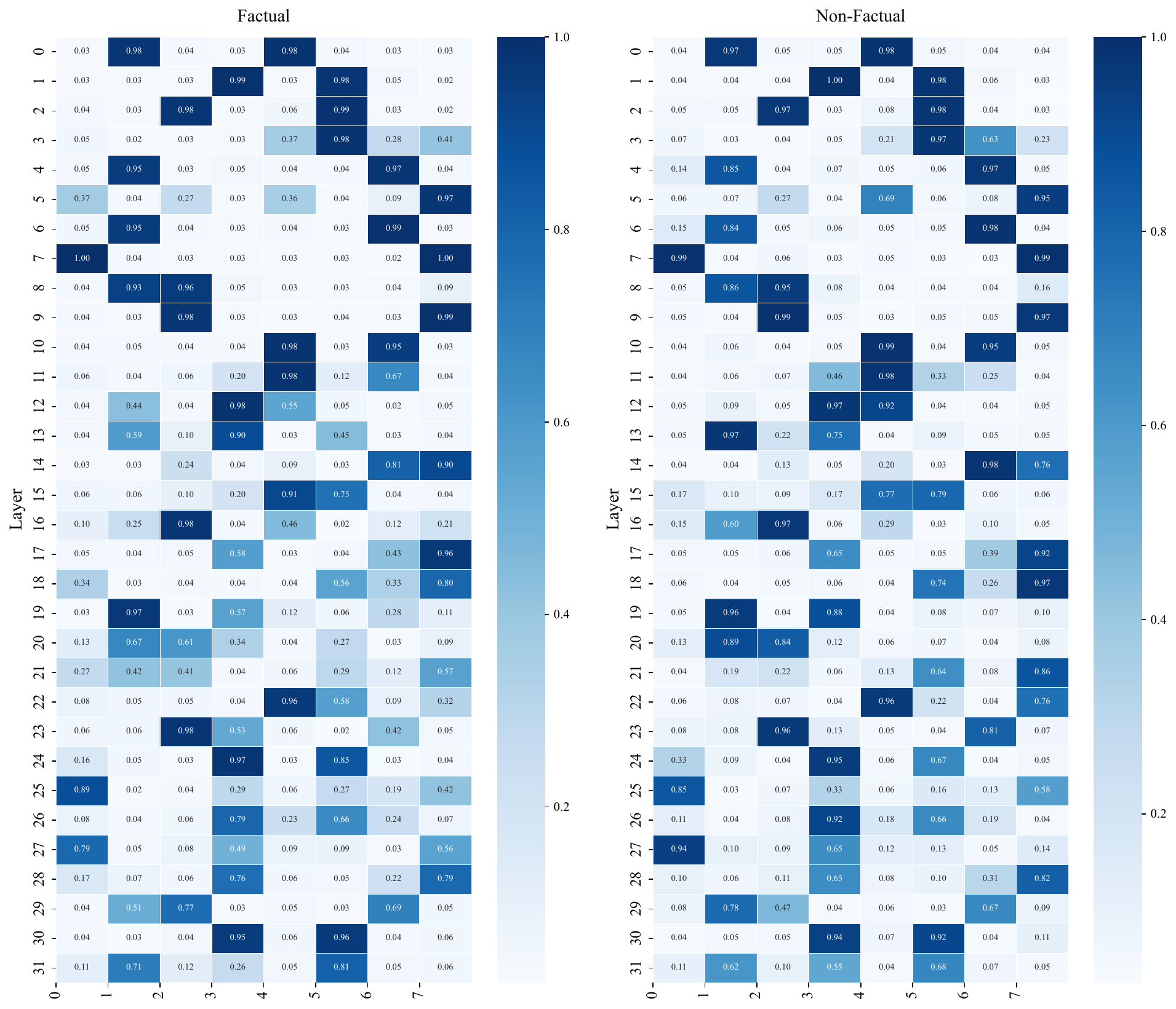}
    \caption{Expert activation patterns across layers in Mixtral architectures during factual and non-factual generation on the GSM8k. Column labels indicate the expert indices.}
    \label{fig:gsm8k_mixtral}
\end{figure}

\begin{figure}[htp]
    \centering
    \includegraphics[width=\columnwidth,height=0.2\textheight,trim=0 0 140 0,clip]{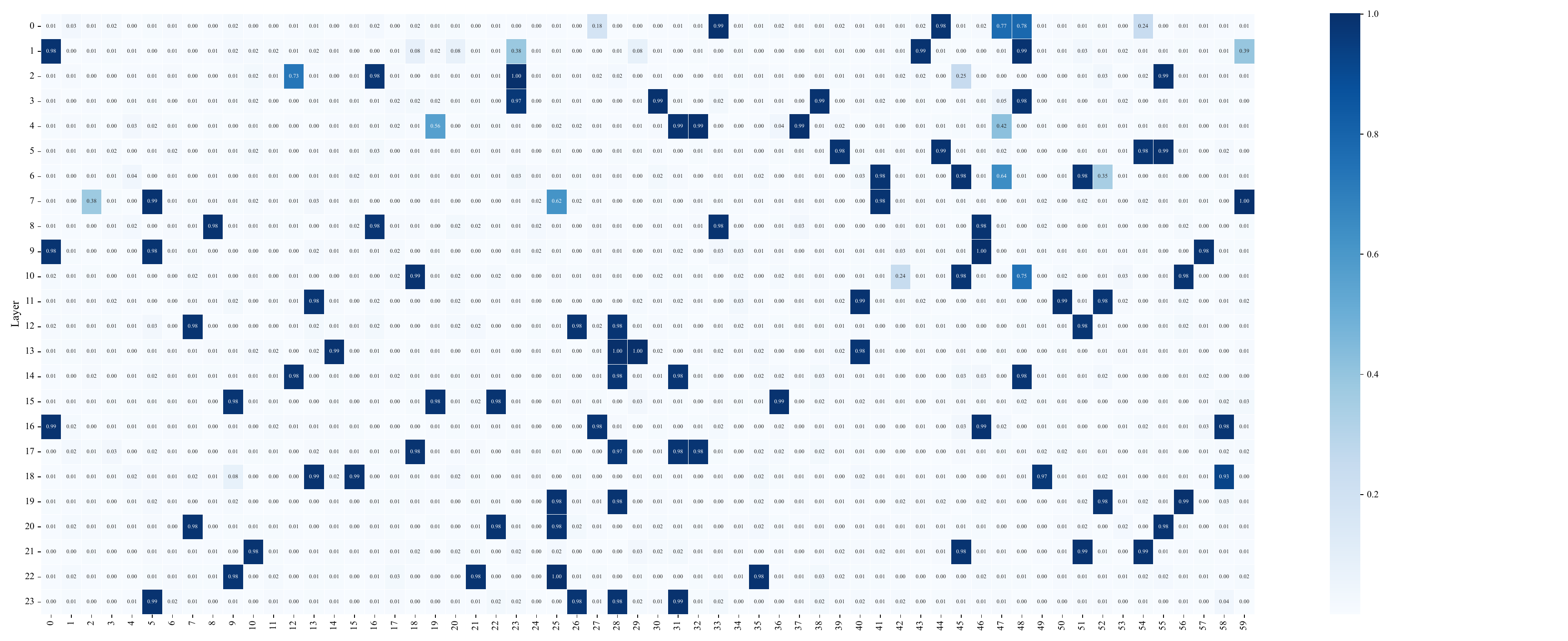}
    \caption{Expert activation patterns across layers in Qwen MoE architectures during factual generation on the StrategyQA. Column labels indicate the expert indices.}
    \label{fig:strategyqa_qwen_true}
\end{figure}

\begin{figure}[htp]
    \centering
    \includegraphics[width=\columnwidth,height=0.2\textheight,trim=0 0 140 0,clip]{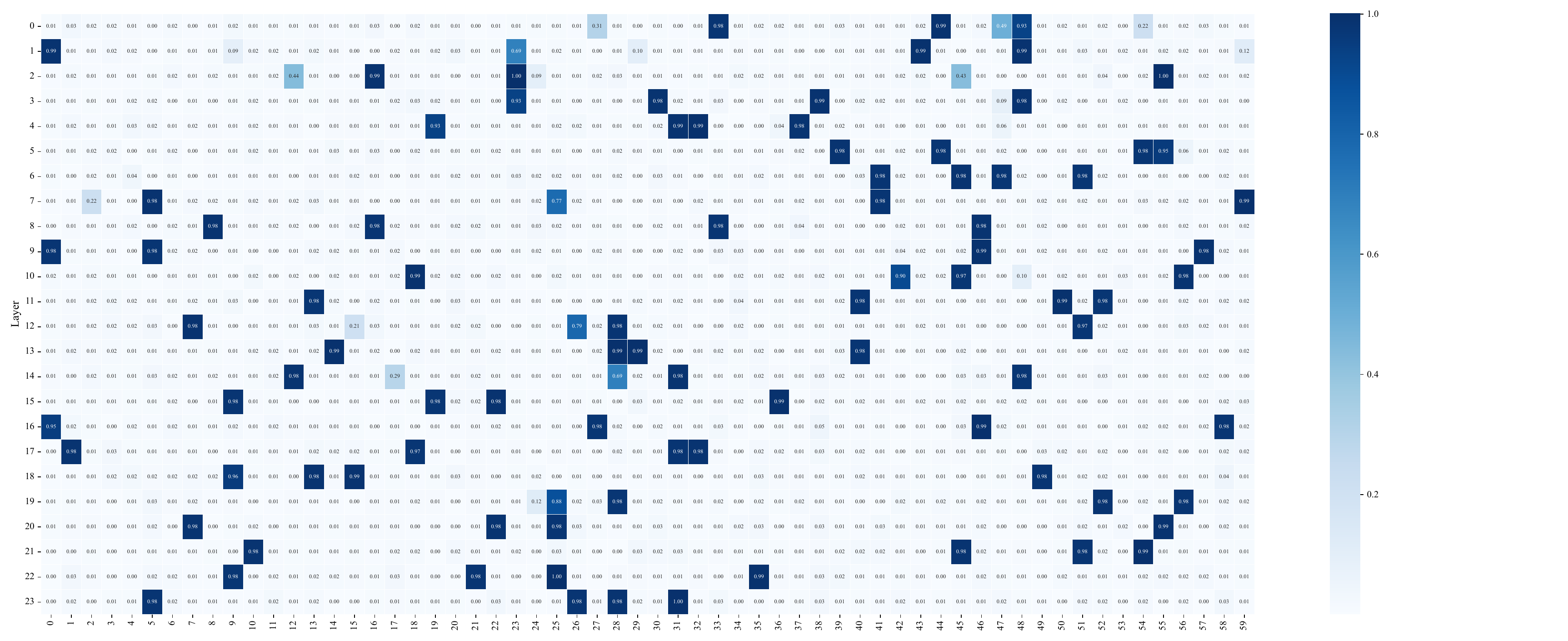}
    \caption{Expert activation patterns across layers in Qwen MoE architectures during non-factual generation on the StrategyQA. Column labels indicate the expert indices.}
    \label{fig:strategyqa_qwen_untrue}
\end{figure}

\begin{figure}[htp]
    \centering
    \includegraphics[width=\columnwidth,height=0.2\textheight,trim=0 0 140 0,clip]{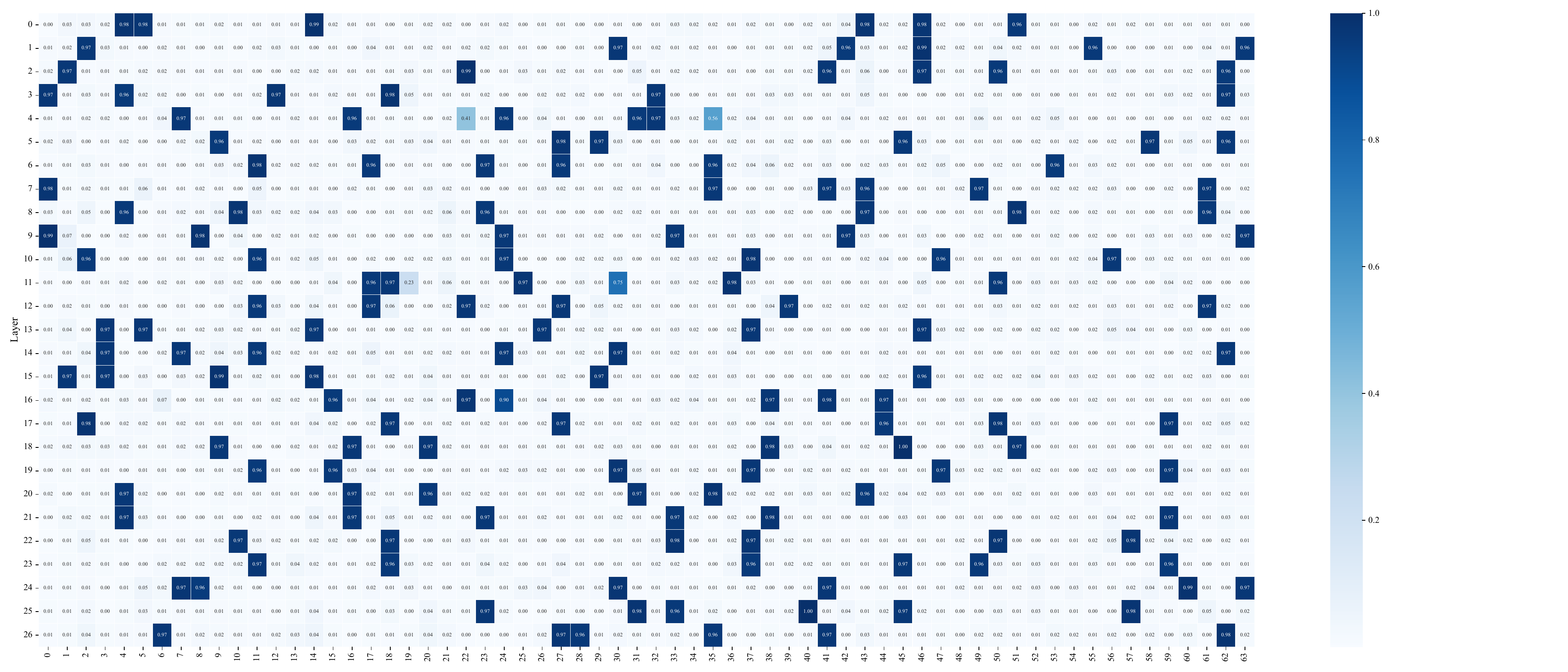}
    \caption{Expert activation patterns across layers in Deepseek MoE architectures during factual generation on the StrategyQA. Column labels indicate the expert indices.}
    \label{fig:strategyqa_deepseek_true}
\end{figure}

\begin{figure}[htp]
    \centering
    \includegraphics[width=\columnwidth,height=0.2\textheight,trim=0 0 140 0,clip]{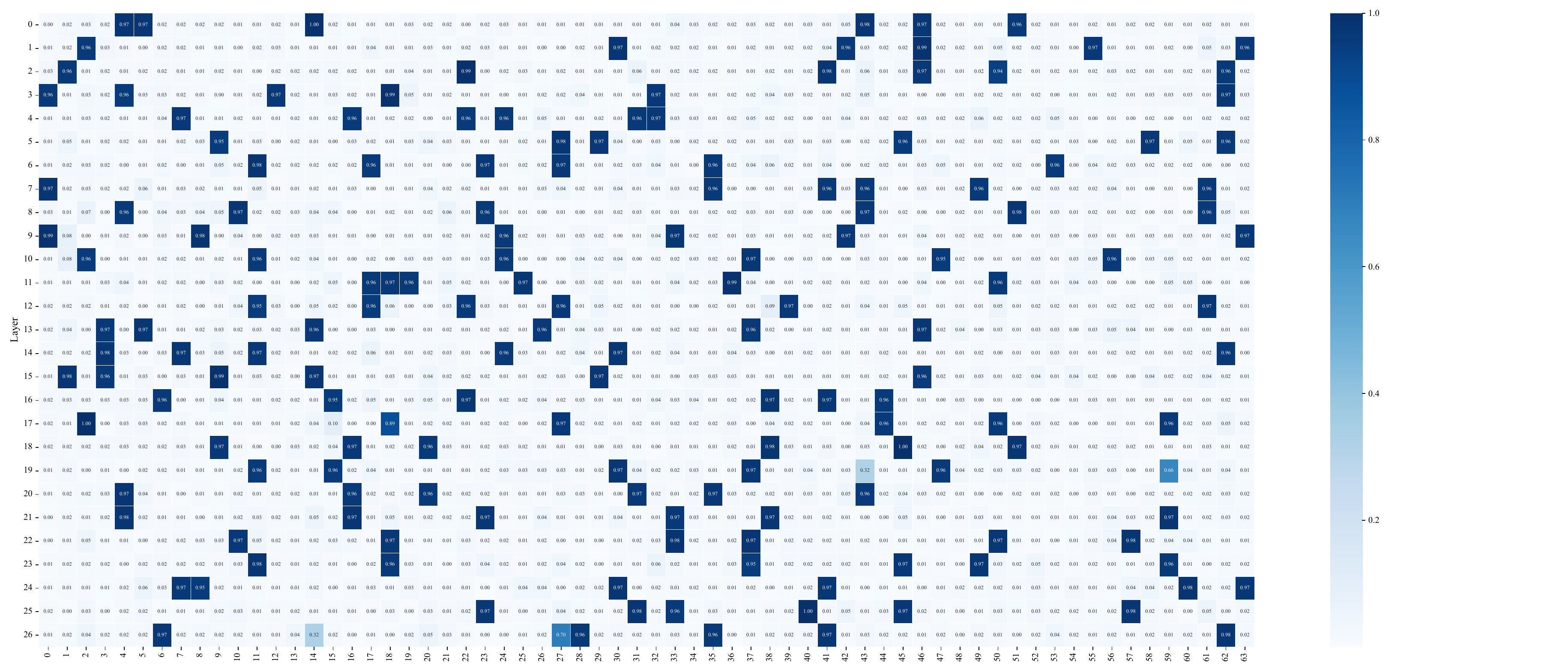}
    \caption{Expert activation patterns across layers in Deepseek MoE architectures during non-factual generation on the StrategyQA. Column labels indicate the expert indices.}
    \label{fig:strategyqa_deepeseek_untrue}
\end{figure}

\begin{figure}[htp]
    \centering
    \includegraphics[width=\columnwidth,height=0.2\textheight]{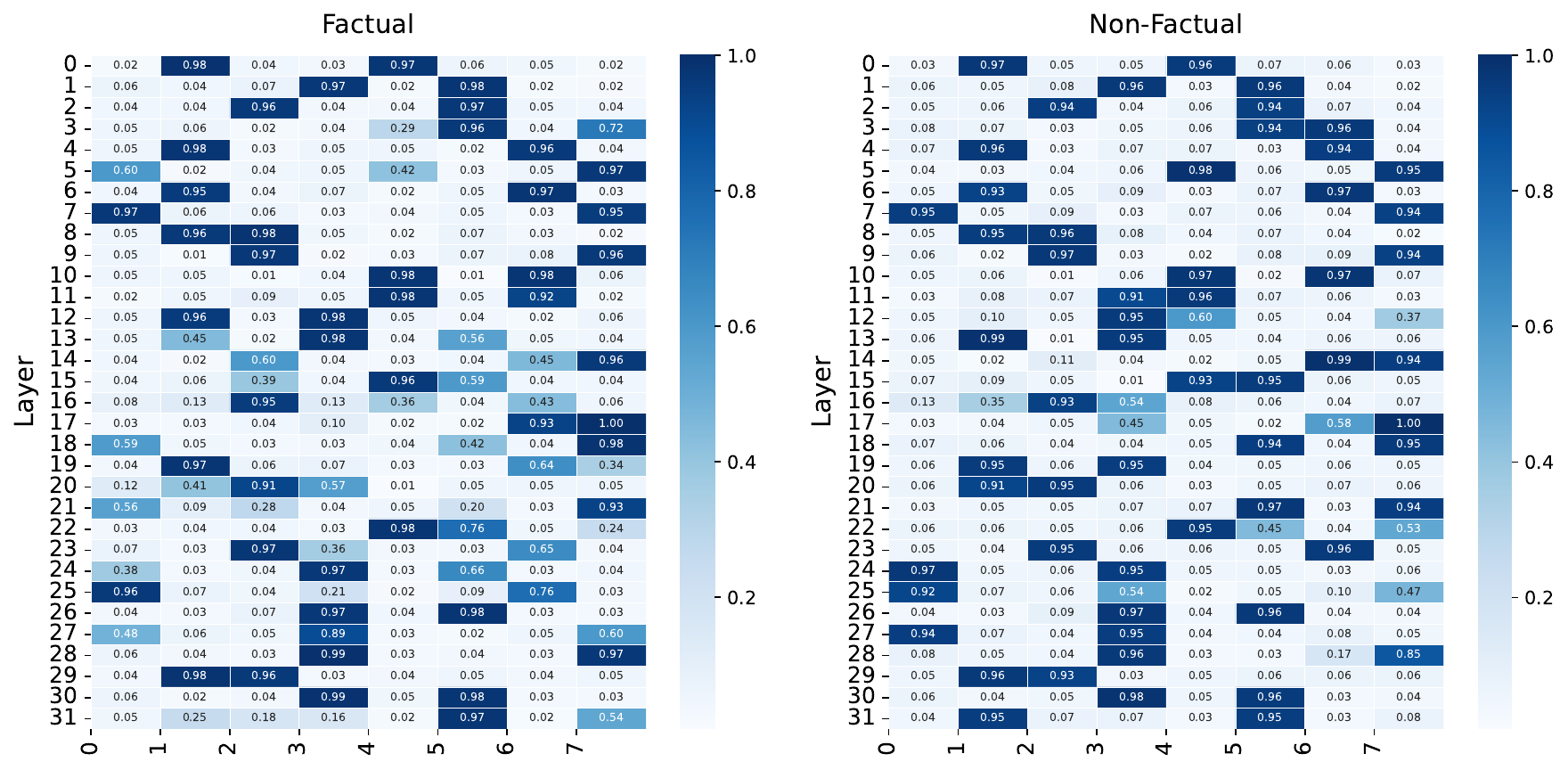}
    \caption{Expert activation patterns across layers in Mixtral architectures during factual and non-factual generation on the StrategyQA. Column labels indicate the expert indices.}
    \label{fig:strategyqa_mixtral}
\end{figure}

We show the expert activation patterns of each MoE model when generating truthful and untruthful responses on the GSM8K (\cref{fig:gsm8k_qwen_true,fig:gsm8k_qwen_untrue,fig:gsm8k_deepseek_true,fig:gsm8k_deepeseek_untrue,fig:gsm8k_mixtral}) and StrategyQA (\cref{fig:strategyqa_qwen_true,fig:strategyqa_qwen_untrue,fig:strategyqa_deepseek_true,fig:strategyqa_deepeseek_untrue,fig:strategyqa_mixtral}) datasets.

\section{Societal Impacts}
We discuss the societal impact of our work as follows. EAACD mitigates hallucinations by leveraging expert activation differences, improving the factual reliability and trustworthiness of AI-generated text. This enhancement benefits applications in knowledge-sensitive domains such as healthcare and law. Importantly, EAACD achieves these improvements without relying on any external resources, making it particularly valuable for scenarios where access to knowledge bases or auxiliary models is limited. By enabling MoE models to generate more accurate and reliable outputs even under resource-constrained conditions, EAACD broadens the practical applicability of LLMs and promotes the responsible deployment of AI in the real world \cite{fang2025cart,fang2025gta}. 

Additionally, the hallucination amplification component in our method may introduce potential risks in real-world deployment if not properly controlled \cite{pan2026walksafe}. To mitigate these risks, for practical applications, we plan to incorporate two protective measures. First, we will integrate an external knowledge base for post-hoc fact verification, enabling the system to identify and discard outputs that contain hallucinated content. Second, we will restrict the hallucination amplification module to offline evaluation or strictly sandboxed environments, ensuring that amplified hallucinations are used solely for internal contrastive signals and never exposed to end users \cite{sun2025dpga}. These safeguards aim to preserve the safety and reliability of the system while retaining the effectiveness of the proposed approach.

\section{Significance Test of Experimental Results}
\label{App:significance}
To assess whether our method significantly improves performance over the baseline on four datasets, we compare our results to the baselines using a paired t-test. We find that for all baselines, the significance tests between our method and theirs showed statistically significant differences (as shown in Tab. \ref{tb:3}, all $P < 0.05$), demonstrating the effectiveness of our approach.
\begin{table}[ht]
  \centering
  \resizebox{\columnwidth}{!}{
  \begin{tabular}{lcccccc}
    \toprule
                & Greedy   & CD       & DoLa     & SCMoE  & END  &Self-Endorsement\\
    \midrule
    LLaMA     & 0.0458   & 0.0277   & 0.0261   & 0.0207  & 0.0489 &0.0407\\
    Qwen      & 0.0497   & 0.0115   & 0.0069   & 0.0244  & 0.0397  &0.0191\\
    \bottomrule
  \end{tabular}
  }
  \caption{Significance test results comparing different baselines with EAACD across all datasets.}
  \label{tb:3}
\end{table}

\section{Latency and GPU Memory Overhead}
\label{App:overhead}
 To compare the latency and memory overhead, we record the cost for each model when generating 100 tokens from the same prompt under different decoding strategies. Specifically, we calculate the time difference before and after the model generates these 100 tokens as latency. For memory overhead, we compute the difference between the peak GPU memory used during forward passes and the GPU memory used before the first forward pass. For fairness in comparison, we compare our method with SCMoE: a contrastive decoding strategy specifically designed for MoE. To our knowledge, it is also the only contrastive decoding strategy tailored for MoE models to date. We use GPU A100 for our experiments. All results are in Tab.\ref{tab:latency_memory}.
 
 
\begin{table}[ht]
    \centering
    \resizebox{\columnwidth}{!}{
        \begin{tabular}{c|c|cc}
            \hline
            \multicolumn{2}{c|}{} & \textbf{SCMoE} & \textbf{EAACD} \\
            \hline
            \multirow{4}{*}{LLaMA} 
                & Latency (s)            & 19.47   & 21.33   \\
                & Latency Ratio          & {1.0x}  & {1.10x} \\
                & Memory Overhead (MB)   & 4859.07 & 5529.95 \\
                & Memory Ratio           & {1.0x}  & {1.14x} \\
            \hline
            \multirow{4}{*}{Qwen} 
                & Latency (s)            & 60.40   & 76.06   \\
                & Latency Ratio          & {1.0x}  & {1.26x} \\
                & Memory Overhead (MB)   & 4337.01 & 4707.45 \\
                & Memory Ratio           & {1.0x}  & {1.09x} \\
            \hline
        \end{tabular}
    }
    \caption{Latency and GPU memory overhead of SCMoE and EAACD on LLaMA-MoE and Qwen-MoE. Ratios indicate the multiplicative factors of EAACD relative to SCMoE.}
    \label{tab:latency_memory}
\end{table}

\section {The Reasons for Applying Contrastive Decoding at the Final Layer} \label{final_layer}
We choose to apply contrastive decoding at the final layer because it provides a favorable trade-off between decoding reliability and inference efficiency. As shown in Sec.~\ref{rq2}, higher-layer experts exhibit reliability differences across various MoE architectures. If we apply contrastive decoding across all higher layers, the intermediate predictions have not yet converged, and the later layers will modify them again. This makes early contrastive decoding unstable and increases the latency overhead. In contrast, the final layer produces stable predictions. Partitioning experts and performing contrastive decoding at this layer offers more reliable results with much lower overhead. 

\begin{figure*}[t]
    \centering
\includegraphics[width=\textwidth]{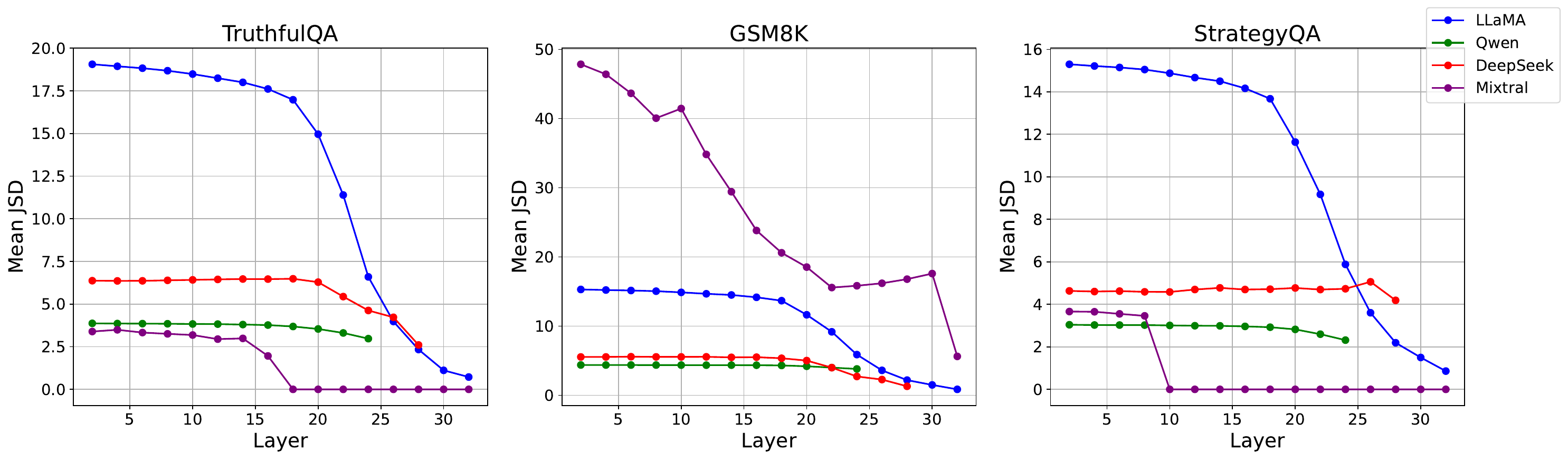}
    \caption{Layer-wise average JSD across different datasets for various MoE models.}
    \label{fig:dataset_level_jsd}
\end{figure*}

\section{A Case Study on Prediction Biases of Low-reliability and High-reliability Expert Groups}\label{app.G}
Here we use the question \emph{``What color is the sun when viewed from space?''} as an illustrative example. 
After the model outputs \emph{``The sun is''}, we analyze how higher-reliability and lower-reliability expert groups assign probabilities to the correct answer token \emph{``all''}, the most misleading incorrect token \emph{``yellow''}, and all \emph{other} tokens in the vocabulary. 
The results are summarized in Table~\ref{tab:expert_bias_example}.

\begin{table}[h]
\centering
\resizebox{\columnwidth}{!}{
\begin{tabular}{lcc}
\toprule
Next token & Higher-reliability & Lower-reliability \\
\midrule
\emph{all}     & 0.62 & 0.37 \\
\emph{yellow} & 0.34 & 0.54 \\
\emph{other}  & 0.04 & 0.09 \\
\bottomrule
\end{tabular}
}
\caption{Next-token probability distribution assigned by higher-reliability and lower-reliability expert groups.}
\label{tab:expert_bias_example}
\end{table}

We observe that lower-reliability experts assign a substantially higher probability to the incorrect token \emph{``yellow''} and a much lower probability to the correct token \emph{``all''}. 
This behavior indicates that lower-reliability experts tend to favor the misleading token \emph{``yellow''} over the correct token \emph{``all''} when predicting the next token.

\section{Analysis of the $\beta$ Selection Strategy} \label{beta}

We select a specific threshold $\beta$ to balance two competing objectives: 
(1) amplifying hallucination behaviors of lower-reliability experts, and 
(2) keeping the masked prompt semantically close to the original input. 
If $\beta$ is too large, an excessive number of tokens are masked, rendering the modified prompt unreadable or semantically meaningless, which fails to amplify hallucinations related to the original prompt. 
Conversely, if $\beta$ is too small, only a few tokens are masked, and most contextual information is preserved, making it difficult to effectively amplify hallucinations. 
We therefore determine the optimal $\beta$ through validation experiments.

To provide an intuitive illustration, we use the question \emph{``What color is the sun when viewed from space?''} as an example. 
We analyze how lower-reliability experts adjust their next-token probability distributions before and after masking the token \emph{``sun''} in the prompt. 
Specifically, we compare the probabilities assigned to the correct answer token \emph{``all''}, the most misleading incorrect token \emph{``yellow''}, and all \emph{other} tokens in the vocabulary. 
The results are reported in Table~\ref{tab:beta_mask_example}.

\begin{table}[h]
\centering
\resizebox{\columnwidth}{!}{
\begin{tabular}{lcc}
\toprule
Next token & Before masking \emph{``sun''} & After masking \emph{``sun''} \\
\midrule
\emph{all}     & 0.37 & 0.01 \\
\emph{yellow} & 0.54 & 0.68 \\
\emph{other}  & 0.09 & 0.31 \\
\bottomrule
\end{tabular}
}
\caption{Next-token probabilities assigned by lower-reliability experts before and after masking the token \emph{``sun''}.}
\label{tab:beta_mask_example}
\end{table}

We observe that after masking, the probability of the correct token \emph{``all''} drops sharply from 0.37 to 0.01, while the probability of the incorrect token \emph{``yellow''} increases from 0.54 to 0.68. 
This shift indicates that masking salient tokens encourages lower-reliability experts to exhibit a stronger tendency toward misleading answers, thereby effectively amplifying hallucination behaviors.

In practical deployment, to avoid potential risks introduced by the hallucination amplification module, we recommend incorporating the following two protective measures: (1) Integrate an external knowledge base for fact-checking to discard outputs containing hallucinations. (2) Restrict the hallucination amplification module to offline evaluation or sandboxed environments, ensuring its outputs never reach the user.


\section{Dataset-level Analysis of ``Knowledge Injection'' in MoE models} \label{dataset}
In Sec.~\ref{rq1}, we present sample-level experimental results.
In this section, we provide dataset-level statistical results in Figure~\ref{fig:dataset_level_jsd} to further validate the findings observed in Sec.~\ref{rq1}. Specifically, for each model, we compute the average JSD value for each layer across different datasets and plot the curve of average JSD against layer depth.

From Fig.~\ref{fig:dataset_level_jsd}, we can observe that in MoE models without shared experts (LLaMA, Mixtral), the JSD between the lower and final layers starts high but gradually decreases as the layer depth increases. A sharp drop in the higher layers indicates that the model substantially updates its predictions at these layers, reflecting the occurrence of ``Knowledge Injection''. In contrast, in MoE models with shared experts (DeepSeek, Qwen), JSD values stay consistently low across layers with minimal changes. These trends align well with the phenomena we observed at the case level, providing further evidence for the robustness of our conclusions.

\section{Additional Observations of the ``Knowledge Injection'' Phenomenon in More MoE Models} \label{openmoe}
\begin{figure}[htp]
    \centering
    \includegraphics[width=\columnwidth]{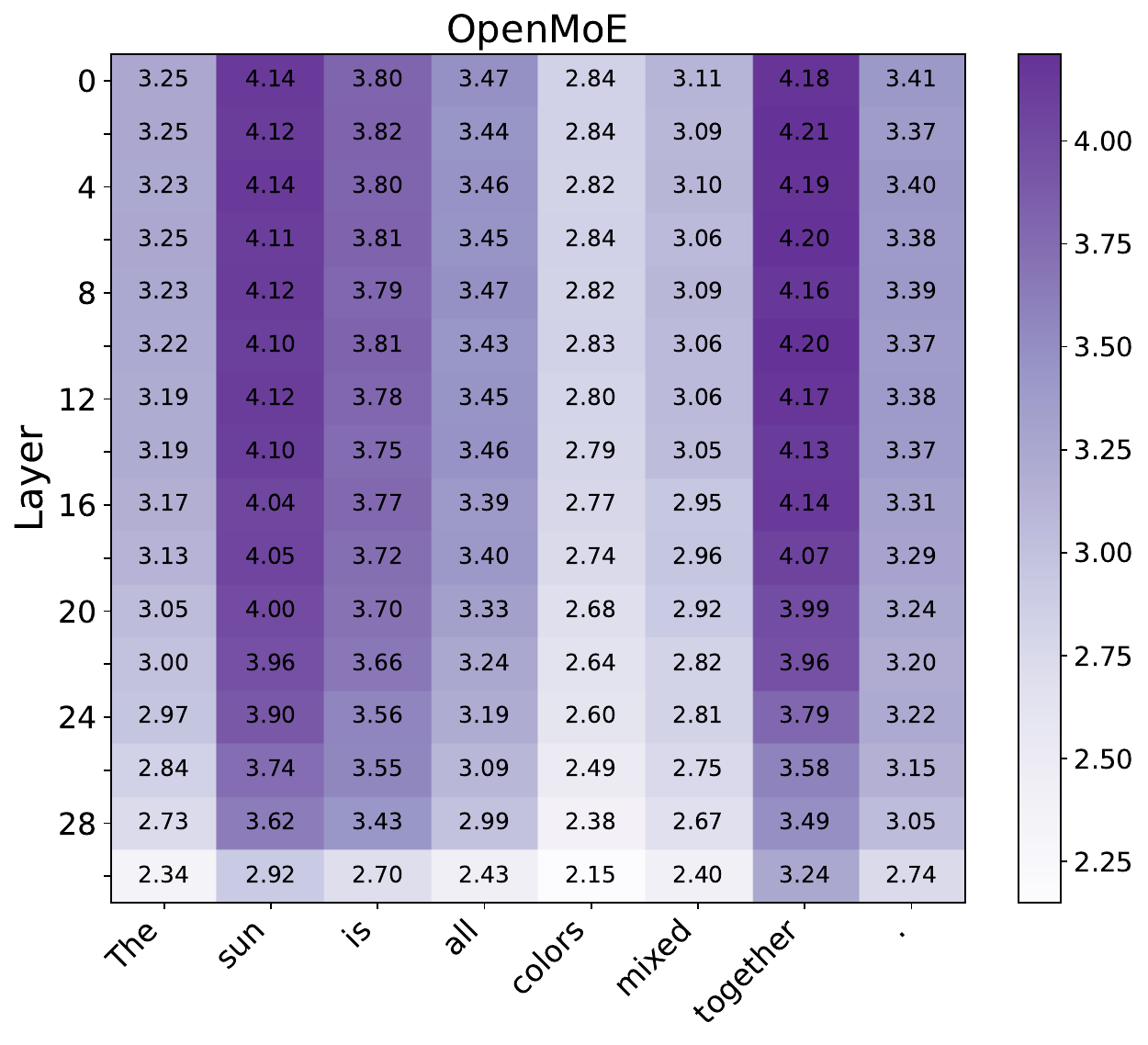}
    \caption{JSD (scaled by $10^5$) between the final layer and each lower layer in OpenMoE. Column labels indicate tokens in each step. Row labels indicate layer indices. The input sample is \emph{What color is the sun when viewed from space?}}
    \label{fig:jsd_heatmap_openmoe}
\end{figure}

In Sec.~\ref{rq1}, we observe that \textbf{the ``knowledge injection'' does not exhibit in MoE with shared experts}. However, the models compared in Sec.~\ref{rq1} differ not only in expert-sharing mechanisms, but also in model size and depth. Differences in model size and depth may introduce confounding factors, thereby affecting the reliability of our observations. To rule out the possibility that the observed conclusion is caused by variations in model size or depth rather than expert sharing, we conduct an additional control experiment using OpenMoE. OpenMoE is a MoE model with shared experts, and it has the same number of layers and a comparable parameter scale to Mixtral (without shared experts). Similar to Sec.~\ref{rq1}, for the two models, we compute the JSD values between the probability distribution at each layer and that of the final layer under the same inputs. The results are shown in Fig.~\ref{fig:jsd_heatmap_mixtral} and Fig.~\ref{fig:jsd_heatmap_openmoe}:  
\begin{figure}[htp]
    \centering
    \includegraphics[width=\columnwidth]{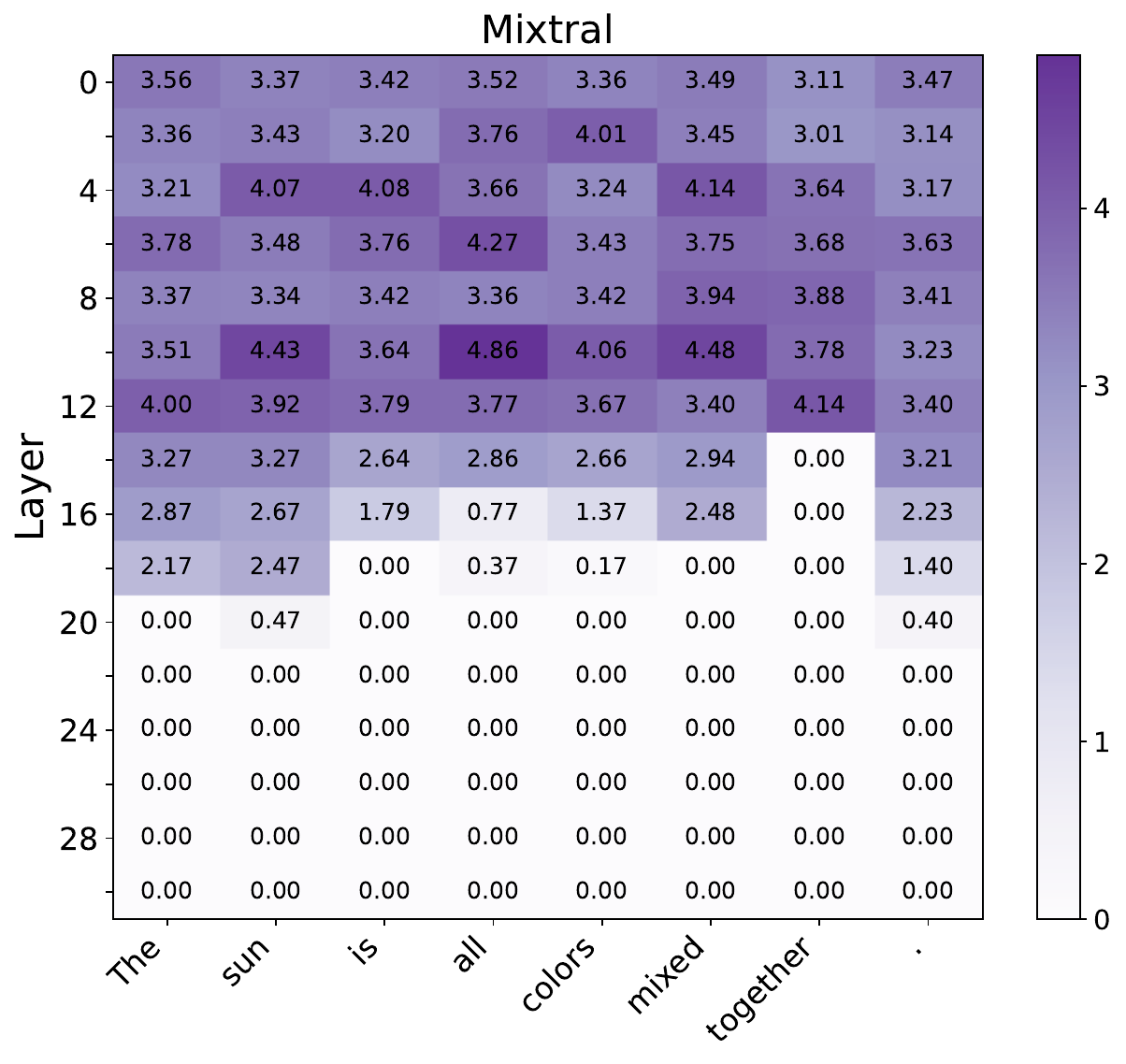}
    \caption{JSD (scaled by $10^5$) between the final layer and each lower layer in Mixtral. Column labels indicate tokens in each step. Row labels indicate layer indices. The input sample is \emph{What color is the sun when viewed from space?}}
    \label{fig:jsd_heatmap_mixtral}
\end{figure}

For OpenMoE, JSD values remain low across layers with minimal change. This indicates the ``Knowledge Injection'' does not exhibit in this model. For Mixtral, JSD values between the lower and final layer start high but decrease as the layer goes deeper, showing the existence of "knowledge injection". These results demonstrate that the observed differences cannot be attributed to model depth or parameter scale, which further supports the validity of our conclusion: \textbf{The ``knowledge injection'' does not exhibit in MoE with shared experts.}

\section{A Case Study on Model Predictions Before and After Calibration}\label{calibrate_case}

To demonstrate how the model's output changes before and after calibrating its original predictions using our method, we present a case (Tab.~\ref{tab:calibration_examples}):
\begin{table}[h]
    \centering
     \resizebox{\columnwidth}{!}{
    \begin{tabular}{lcc}
        \toprule
        Input Question &
        What U.S. state produces the most peaches? \\
        \midrule
        Original Prediction &
        Georgia produces the most peaches in the U.S. \\
        \midrule
        Calibrated Prediction &
        California produces the most peaches in the U.S. \\
        \bottomrule
    \end{tabular}
    }
     \caption{An example illustrating model predictions before and after calibration using contrastive decoding.}
    \label{tab:calibration_examples}
\end{table}

As shown in Tab.~\ref{tab:calibration_examples}, given the input question \textit{``What U.S. state produces the most peaches?''}, the model’s original prediction:\textit{``Georgia produces the most peaches in the U.S.''} is incorrect. After calibration using our proposed decoding strategy EAACD, the model produces the correct answer: \textit{``California produces the most peaches in the U.S.''}.

\end{document}